\documentclass[lettersize,journal]{IEEEtran}
\usepackage{amsmath,amsfonts}
\usepackage[dvipsnames]{xcolor}
\usepackage{algorithmic}
\usepackage{algorithm}
\usepackage{array}
\usepackage{cite}
\usepackage{siunitx}
\usepackage{textcomp}
\usepackage{stfloats}
\usepackage{tcolorbox}  
\usepackage{url}
\usepackage{booktabs}
\usepackage{cite}
\usepackage{multirow}
\usepackage{verbatim}
\usepackage{adjustbox}
\usepackage{graphicx}
\usepackage{booktabs}
\usepackage{subfigure}
\usepackage{subcaption}
\usepackage{color}
\usepackage{hyperref} 
\newcommand{\mytheoremname}{\bfseries Theorem}
\newcommand{\mylemmaname}{\bfseries Lemma}
\newcommand{\myproof}{\bfseries Proof}
\newcommand{\mypropositionname}{\bfseries Proposition}
\newcommand{\mydefinitionname}{\bfseries Definition}
\newtheorem{Theorem}{\mytheoremname} 
\newtheorem{Lemma}{\mylemmaname}
\newtheorem{Proof}{\myproof}
\newtheorem{Definition}{\mydefinitionname} 
\tcolorboxenvironment{Theorem}{
  colback=Blue!0.5!White,  
  colframe=white, 
  boxrule=0.1pt,         
}
\tcolorboxenvironment{Lemma}{
  colback=Blue!0.5!White,  
  colframe=white, 
  boxrule=0.1pt,         
}

\begin{document}

\title{Towards Trustworthy Hypergraph Neural Networks under Label Noise}
\author{Mengyao Zhou, Zhiheng Zhou, Xiao Han,  Guiying Yan
\thanks{ This work was supported by the National Natural Science Foundation of China (No. 12231018)and the Postdoctoral Innovation Program of Shandong Province (No.SDCX-ZG-202603012).(Corresponding authors:  Guiying Yan.)}
\thanks{Mengyao Zhou and Guiying Yan are with the Academy of Mathematics and Systems Science, Chinese Academy of Sciences and also with the University of Chinese Academy of Sciences, Beijing 100190, China (e-mail: zhoumengyao@amss.ac.cn; yangy@amss.ac.cn).}
\thanks{Zhiheng Zhou is with the School of Mathematics and Statistics, Shandong University, Weihai, Shandong 264209, China (e-mail: zhouzhiheng@amss.ac.cn).}
\thanks{Xiao Han is with School of Artificial Intelligence, Beihang University, Beijing 100191,China (e-mail: hx2210@buaa.edu.cn).}
}



\maketitle
This work has been submitted to the IEEE for possible publication. Copyright may be transferred without notice, after which this version may no longer be accessible.

\begin{abstract}
Hypergraph neural networks (HGNNs) have demonstrated remarkable capabilities in processing complex higher-order relationships. However, their performance is highly dependent on labeled data, making them vulnerable to label noise. Despite advances in learning with label noise (LLN) and graph learning with label noise (GLN), noisy-label learning on hypergraphs remains underexplored. In this paper, we present a systematic study of hypergraph node classification under label noise. First, we adapt representative LLN and GLN methods to hypergraphs and evaluate them under a unified benchmark, revealing the limitations of existing robust learning strategies for hypergraphs. Building on this, we propose a new hypergraph robust framework, HyperTrust, which first estimates hyperedge trustworthiness through a pretraining-based, entropy-aware strategy, and then incorporates the HyperedgeBoost module to enhance reliable supervision by connecting unlabeled nodes to trustworthy hyperedges, as well as the HyperedgePrune module to suppress noisy propagation by removing untrustworthy node–hyperedge incidences. Finally, two modules work collaboratively to  adjust the hypergraph structure  and generate final predictions. Extensive experiments and theoretical analysis demonstrate the effectiveness and robustness of HyperTrust on multiple hypergraph datasets under various noisy settings. Our work provides a unified benchmark and an effective solution for hypergraph learning with label noise and lays a foundation for future research in this direction. 
\end{abstract}

\begin{IEEEkeywords}
hypergraph, hypergraph neural network, label noise,  benchmark, robustness.
\end{IEEEkeywords}

\section{Introduction}
\IEEEPARstart{I}{n}  recent years, hypergraphs have attracted widespread attention as an important tool for modeling complex systems~\cite{zhou2006learning,antelmi2023survey}. Unlike traditional graphs, which can only represent pairwise relationships between nodes, hypergraphs can connect multiple nodes through hyperedges, naturally capturing higher-order relations~\cite{berge1984hypergraphs}. This capability gives hypergraphs unique expressive power in many domains such as social networks~\cite{zlatic2009hypergraph,zhu2018social}, recommendation systems~\cite{xia2022self,wang2020next}, and biological networks~\cite{feng2021hypergraph,klamt2009hypergraphs}. To fully exploit the information from hypergraph data, researchers have proposed various hypergraph neural networks (HGNNs)~\cite{kim2024survey}, achieving significant progress in tasks such as node classification~\cite{wu2022hypergraph}, link prediction~\cite{yadati2020nhp,li2013link}, and representation learning~\cite{antelmi2023survey,wang2026hypersynergyx}.\\
Despite these successes, existing HGNNs heavily rely on labeled data~\cite{zhou2025graph}. However, in real-world scenarios, labels are often imperfect due to annotation ambiguity, insufficient expert knowledge, crowdsourcing errors, or automatic labeling pipelines~\cite{frenay2013classification}. Once noisy labels are introduced into the training set, HGNNs may easily overfit these noisy supervision signals, leading to degraded generalization ability and poor robustness. More importantly, compared with graphs, label noise in hypergraphs can be even more harmful~\cite{dang2021noise}. Since hyperedges usually involve multiple nodes, a noisy node may influence a larger set of neighbors through high-order message passing, thereby amplifying the propagation of erroneous information. Therefore, improving the robustness of HGNNs under noisy labels is an important and challenging problem.\\
To address label noise, extensive research has been conducted in recent years. In the general learning with label noise (LLN) literature, representative strategies include sample selection~\cite{Han_Yao_Yu_Niu_Xu_Hu_Tsang_Sugiyama_2018,Yu_Han_Yao_Niu_Tsang_Sugiyama_2019, Malach_Shalev-Shwartz_2017, Li_Socher_Hoi_2020,Jiang_Zhou_Leung_Li_Li_2017}, loss regularization~\cite{goldberger2017training,Ma_Wang_Houle_Zhou_Erfani_Xia_Wijewickrema_Bailey_2018,Patrini2017over,Reed_Lee_Anguelov_Szegedy_Erhan_Rabinovich_2014}. These methods have shown strong effectiveness on Euclidean data such as images. Meanwhile, in graph learning, graph learning with label noise (GLN) methods have been proposed to improve the robustness of graph neural networks by leveraging graph-specific structural information, such as data augmentation~\cite{dai2021nrgnn,qian2023robust}, loss regularization~\cite{li2021unified,du2021noise, zhang2020adversarial} and contrastive learning~\cite{yuan2023learning}. 
However, these methods are primarily designed for Euclidean data or graphs, and their assumptions do not fully match the higher-order dependencies in hypergraphs~\cite{Ghosh_Kumar_Sastry_2022,Jiang_Zhou_Leung_Li_Li_2017}. Moreover, although some existing strategies may be adapted to hypergraph data, their effectiveness and limitations under higher-order relational structures remain unclear. Therefore, hypergraph learning under label noise  is still largely underexplored and lacks a systematic investigation.\\
To bridge this gap,  we present a systematic study of hypergraph node classification under label noise. Specifically, we adapt representative LLN and GLN methods~\cite{wang2024noisygl} to hypergraph and evaluate them under a unified benchmark. This evaluation enables a fair comparison among different robust learning strategies and, to the best of our knowledge, represents one of the first systematic investigations of noisy-label learning on hypergraphs. The empirical results show that existing robust methods do not consistently retain their effectiveness when adapted to hypergraphs. These observations indicate that noisy-label learning on  hypergraph may not be adequately addressed by directly reusing existing graph-oriented or generic noisy-label techniques, and instead requires a  hypergraph-specific design.\\
Building on these findings, we further propose a new robust hypergraph framework, named HyperTrust. The core idea of HyperTrust is to explicitly model hyperedge trustworthiness, thereby promoting trustworthy  supervision propagation while suppressing the spread of noisy information. Specifically, HyperTrust first estimates hyperedge trustworthiness in an entropy-aware manner, distinguishing trustworthy and untrustworthy  hyperedges according to the label consistency of their incident nodes. Based on this estimation, we design two complementary modules. The HyperedgeBoost module enhances reliable  supervision by connecting unlabeled nodes to trustworthy hyperedges according to node--hyperedge similarity, thereby facilitating reliable higher-order information propagation. In contrast, the HyperedgePrune module suppresses noisy propagation by removing unreliable node--hyperedge incidences from untrustworthy hyperedges, thus reducing the interference caused by corrupted supervision. Finally, the two modules collaborate to  adjust the hypergraph structure and generate the final predictions.\\
Through theoretical analysis, we show that HyperTrust can improve the expected classification margin of hypergraph neural networks under label noise, providing a theoretical explanation for its robustness. Empirically, we conduct extensive experiments on multiple hypergraph datasets under various types of label noise, and further evaluate the proposed method with different hypergraph backbones. The results consistently demonstrate that HyperTrust achieves superior  performance against  baselines across a wide range of settings, validating its robustness and effectiveness for noisy-label hypergraph node classification.
The main contributions of this paper can be summarized as follows:
\begin{itemize}
    \item We present a systematic study of hypergraph node classification under label noise by adapting representative LLN and GLN methods to hypergraphs and evaluating them under a unified benchmark, which reveals the limitations of existing robust learning strategies in the hypergraph setting.
    \item We propose a novel hypergraph-specific robust framework, HyperTrust, which estimates hyperedge trustworthiness in an entropy-aware manner and jointly incorporates HyperedgeBoost and HyperedgePrune to enhance reliable supervision and suppress noisy propagation.
    \item Through theoretical analysis, we show that HyperTrust improves the expected classification margin of hypergraph neural networks under label noise, and extensive experiments on multiple hypergraph datasets with different backbones further verify its effectiveness and robustness under diverse noisy settings.
\end{itemize}
\section{Related Work}
In this section, we review literature closely related to our work, focusing on two key areas: Hypergraph Neural Networks (HGNNs) and Learning with Label Noise.
\subsection{Hypergraph Neural Networks}
Hypergraph neural networks (HGNNs) extend  graph neural networks (GNNs) by explicitly modeling high-order relationships among multiple entities, going beyond pairwise interactions. The early development of this field is marked by the HGNN~\cite{feng2019hypergraph}, which adopts a spectral convolution paradigm to aggregate information from nodes within shared hyperedges.

Subsequent research has shifted toward more flexible message-passing formulations. Representative models such as HNHN~\cite{dong2020hnhn}, HyperGCN~\cite{yadati2019hypergcn}, HyperSAGE~\cite{arya2020hypersage}, and UniGNN~\cite{huang2021unignn}  enhance the expressive capacity of HGNNs by explicitly modeling node–hyperedge interactions and designing dedicated aggregation schemes, enabling more effective representation of complex relational structures.

More recent advances further relax the reliance on spectral assumptions. Architectures such as AllDeepSets~\cite{chien2021you} and AllSetTransformer~\cite{chien2021you} treat hyperedges as unordered sets and employ permutation-invariant functions, enabling more flexible and principled modeling of high-order interactions. In addition, emerging directions explore improved training dynamics and long-range dependency modeling, such as ODE-based frameworks~\cite{yan2024hypergraph,zhou2026hypergraph}  and path-aware representations~\cite{xie2025k}, which further expand the capability of hypergraph learning in complex relational settings.

\subsection{Learning with Label Noise}
Deep Neural Networks (DNNs) are prone to overfitting noisy labels, which degrades generalization performance~\cite{Zhang2016over}. To address this issue, existing methods  mainly follow two directions: sample selection\cite{Han_Yao_Yu_Niu_Xu_Hu_Tsang_Sugiyama_2018,Yu_Han_Yao_Niu_Tsang_Sugiyama_2019, Malach_Shalev-Shwartz_2017, Li_Socher_Hoi_2020,Jiang_Zhou_Leung_Li_Li_2017} and loss regularization ~\cite{goldberger2017training,Ma_Wang_Houle_Zhou_Erfani_Xia_Wijewickrema_Bailey_2018,Patrini2017over,Reed_Lee_Anguelov_Szegedy_Erhan_Rabinovich_2014}. Sample selection approaches, such as Co-teaching and Co-teaching+\cite{Yu_Han_Yao_Niu_Tsang_Sugiyama_2019}, identify and reweight clean samples during training, while loss regularization methods (e.g., Backward, Forward)~\cite{goldberger2017training,Ma_Wang_Houle_Zhou_Erfani_Xia_Wijewickrema_Bailey_2018,Patrini2017over,Reed_Lee_Anguelov_Szegedy_Erhan_Rabinovich_2014}, explicitly model or correct label noise within the loss function.

However, these methods are primarily developed for Euclidean data and are not directly applicable to graph scenarios~\cite{Ghosh_Kumar_Sastry_2022},\cite {Jiang_Zhou_Leung_Li_Li_2017}. In Graph Neural Networks (GNNs), label noise is more challenging due to the message-passing mechanism, which propagates incorrect labels across nodes. Existing robust GNN methods typically fall into three categories: data augmentation (e.g., NRGNN~\cite{dai2021nrgnn}.), loss regularization(e.g., CP) and contrastive learning(e.g., CGNN~\cite{li2024contrastive}). Despite their effectiveness, these approaches are primarily designed for  graph structures and are not well-suited for modeling higher-order relationships.

In hypergraph models such as  HGNN~\cite{feng2019hypergraph}, noisy labels can propagate through hyperedges and simultaneously affect multiple nodes, making the problem more challenging. Although some existing noisy-label learning methods can be  extended to hypergraph data, their effectiveness in capturing high-order noise propagation remains unclear. Therefore, noisy-label learning on hypergraphs deserves more dedicated and systematic investigation.

\section{Preliminary}
\subsection{Notations} 
 Let $\mathcal{H} = (\mathcal{V} , \mathcal{E})$ denote a  hypergraph, where $\mathcal{V}$ is the vertex set containing $n$ unique vertices and $\mathcal{E}$ is the edge set containing $m$ hyperedges.  The hypergraph can be represented by an incidence matrix $H\in R^{n\times m}$  where $H_{ij} = 1$ if the vertex ${v}_{i}\in \mathcal{V}$ is contained in the hyperedge $  {e_j} \in \mathcal{E}$, otherwise 0.  Let $\mathbf{X}=[\mathbf{x}_{1},\mathbf{x}_{2},... \mathbf{x}_{n}]^{T} $ denote the node feature matrix, and $\mathbf{x}_{i}$ is associated with the node $v_{i}$. Each hyperedge $e_j \in \mathcal{E}$ is assigned a  weight $w_{e_j}$, all the weights formulate a diagonal matrix $W \in R^{m\times m}$.
 The vertex and edge degree of hypergraph can be expressed  as $d_{i} = \sum_{j=1}^{m} H_{ij}$ and $d_{e_{j}} = \sum_{i=1}^{n} H_{ij}$, all the degrees formulate the diagonal matrix $D_{v}$ and $D_{e}$. 
\subsection{Noisy types}
In real-world scenarios, label noise can be quite complex. In this paper, we focus on the   following most commonly encountered types of noise~\cite{wang2024noisygl},~\cite{zhou2025graph}:\par
\textbf{Pair noise:} This type of noise assumes that the true label can only be flipped to its corresponding pair class with a probability $\epsilon$. Formally, assume that the label set $D=\left\{d_{1},d_{2}...d_{C}\right\}$, for ${\forall}{v
_{j}\in \mathcal{V}_{L}}$, ${\exists!}{s\neq t}$, we have $p(y^{N}_{j} = d_{s}|y^{T}
_{j}= d_{t}) = {\epsilon}$, where $C$ represents the number of classes.\par
\textbf{Uniform noise:} This type of noise assumes that the true label has a probability of $\epsilon \in (0, 1)$ to be uniformly flipped to another class. Formally, assume that the label set $D=\left\{d_{1},d_{2}...d_{c}\right\}$, for ${\forall}{v
_{j}\in \mathcal{V}_{L}}$,${\forall}{s\neq t}$, we have $p(y^{N}_{j} = d_{s}|y^{T}
_{j}= d_{t}) = \frac{\epsilon}{C-1}$, where $C$ represents the number of classes.\par

\textbf{Random noise:} This type of noise assumes that the true label can be flipped to any other class with a probability $\epsilon$, but the probability of flipping to each class is not uniform and may be varied.

\subsection{Hypergraph neural network}
Hypergraph Neural Networks leverage both the Hypergraph structure and node features $\mathbf{X}$ to learn a node's representation vector $h_{v}$, or the entire graph's representation $h_{H}$. Typically, this process consists of two stages: first, node features are aggregated into hyperedge representations; then, hyperedge representations are propagated back to incident nodes. After $k$ iterations, each node representation captures high-order structural information from its connected hyperedges. Formally, the $k\mbox{-}th$ layer of an HGNN is defined as:
\begin{equation}
\begin{aligned}
\mathbf{h}_{e}^{k} &= \mathrm{AGGREGATE}_{\mathrm{node}\rightarrow\mathrm{edge}}^{(k)}
(\{\mathbf{h}_{v}^{k-1}: v \in \mathcal{V}(e)\}),\\
\mathbf{h}_{v}^{k} &= \mathrm{AGGREGATE}_{\mathrm{edge}\rightarrow\mathrm{node}}^{(k)}
(\{\mathbf{h}_{e}^{k}: e \in \mathcal{E}(v)\}),
\label{equation_hgnn}
\end{aligned}
\end{equation}
where $\mathcal{V}(e)$ denotes the set of nodes incident to hyperedge $e$, $\mathcal{E}(v)$ denotes the set of hyperedges connected to node $v$, $\mathbf{h}_{e}^{k}$ is the representation of hyperedge $e$ at the $k\mbox{-}th$ layer, and $\mathbf{h}_{v}^{k}$ is the representation of node $v \in V$ at the $k\mbox{-}th$ layer. The specific forms of aggregate function may vary across different HGNN models.

\subsection{Problem Definition}
The problem of learning a robust HGNN with noise  is formally defined as:
 Given a hypergraph $\mathcal{H} = (\mathcal{V} , \mathcal{E})$  with a  set of nodes $\mathcal{V} _L\subseteq \mathcal{V} $ provided with noisy labels $Y_{L}^{N}$, we aim to learn a robust HGNN which predicts the true labels of the unlabeled nodes in $\mathcal{V} _{U}$. 

\section{Benchmark Design and Empirical Results}
\label{Benchmark Design and Empirical Motivation}
\subsection{Benchmark Design}
\begin{figure}[t]
\centering  
\includegraphics[width=8.5cm,height =6.2cm]{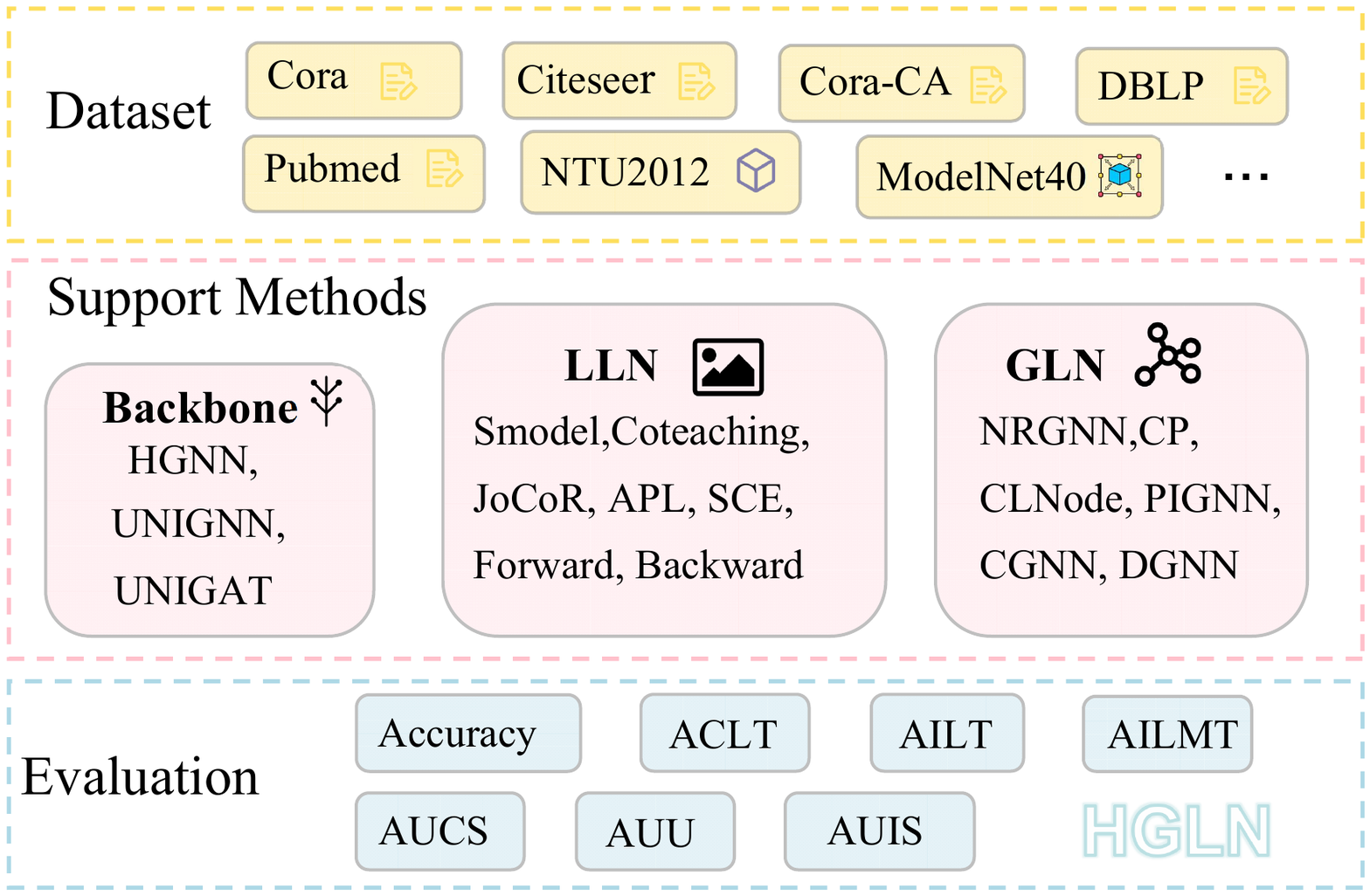}
 \caption{Benchmark design for hypergraph learning under label noise. The benchmark covers diverse datasets from co-citation, co-authorship, and visual-object domains, and includes three categories of methods for comparison: hypergraph backbones, adapted LLN methods, and adapted GLN methods. All methods are evaluated under a unified protocol using multiple metrics, including overall accuracy, ACLT, AILT, AILMT, AUCS, AUU, and AUIS.}
\label{benchmark}
\end{figure}

\textbf{  Datasets, Baselines and Metrics.}
To comprehensively evaluate the effectiveness of different robust learning strategies on hypergraphs, we construct a unified benchmark, as illustrated in Fig.~\ref{benchmark}, which covers diverse datasets from co-citation, co-authorship and visual object, including Cora, Citeseer, Cora-CA, DBLP-CA, Pubmed~\cite{yadati2019hypergcn}, NTU2012 and ModelNet40~\cite{zhou2026tackling}. We include three categories of representative methods in this benchmark. The first category consists of hypergraph backbone models, where HGNN serves as the reference without explicit noise-robust mechanisms. The second category includes LLN methods, such as S-model~\cite{goldberger2017training}, Co-teaching~\cite{Yu_Han_Yao_Niu_Tsang_Sugiyama_2019}, JoCoR~\cite{wei2020combating}, APL~\cite{ma2020normalized}, SCE~\cite{wang2019symmetric}, Forward~\cite{Patrini2017over}, and Backward~\cite{Patrini2017over}, which mainly address label noise through sample selection, label correction, and loss reweighting. The third category contains GLN methods, including NRGNN~\cite{dai2021nrgnn}, CP~\cite{zhang2020adversarial}, CLNode~\cite{wei2023clnode}, PIGNN~\cite{du2021noise}, CGNN~\cite{yuan2023learning}, and DGNN~\cite{nt2019learning}, which improve robustness through techniques such as data augmentation, regularization, and contrastive learning. All methods are evaluated under a unified protocol using multiple metrics~\cite{wang2024noisygl}, including accuracy, the accuracy of correctly labeled (ACLT), incorrectly labeled (AILT), and misleadingly labeled (AILMT) training nodes, accuracy of Unlabeled Correctly Supervised nodes(AUCS), accuracy of Unlabeled Unsupervised nodes(AUU), and accuracy of Unlabeled Incorrectly Supervised nodes(AUIS). By covering representative datasets, methods, and evaluation metrics, our benchmark enables a systematic assessment of the effectiveness and limitations of existing robust learning paradigms in hypergraph settings.

\textbf{Implementations.}
To ensure fair comparisons while preserving the original design principles of each method, we adopt a unified, faithful, and controlled adaptation strategy when extending existing approaches to hypergraphs. First, all methods are implemented with the same HGNN backbone, minimizing performance variations caused by different representation learning architectures and allowing us to focus on the robustness mechanisms themselves. Second, we retain the core components of each method (e.g., sample selection, co-training schemes, and loss correction strategies) and perform minimal modifications only to the representation learning module to accommodate hypergraph structures. Specifically, the original graph convolution or representation learning operations are replaced with their hypergraph counterparts, while keeping the training objectives and optimization procedures unchanged. Furthermore, all methods are trained and evaluated under identical supervision conditions, including the same noisy labels, data splits, and training protocols, without introducing any additional prior knowledge or  annotations. We strictly follow the original papers and publicly available implementations for all methods, and more details are provided in  Appendix~\ref{C}.
\begin{figure*}[htb]
\centering  
\subfigure[Pair(0.3)]
{
\includegraphics[width=5.5cm,height =4.12cm]{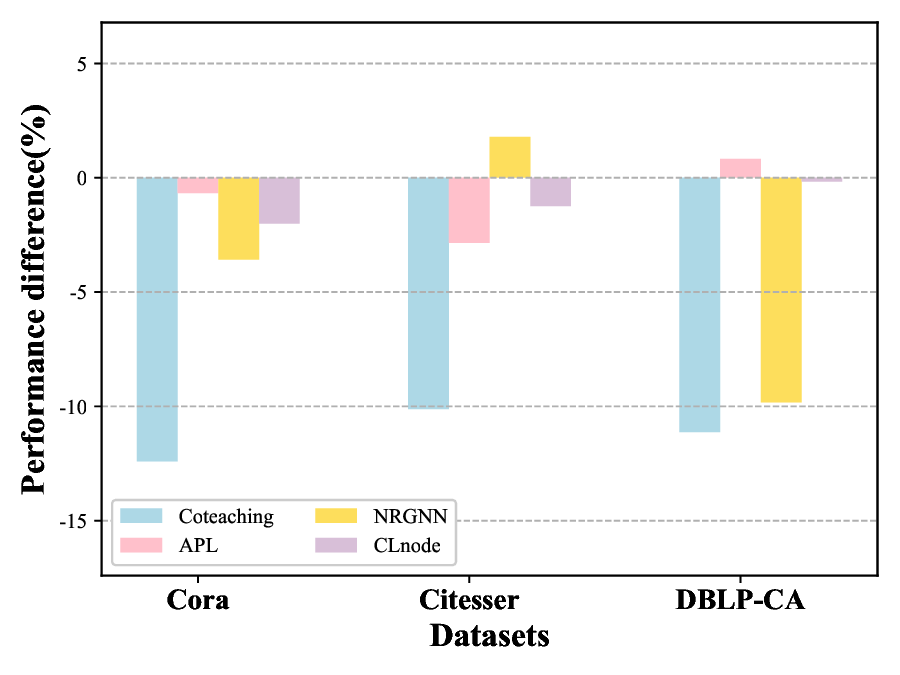}}
\subfigure[Uniform(0.3)]
{
\includegraphics[width=5.5cm,height =4.12cm]{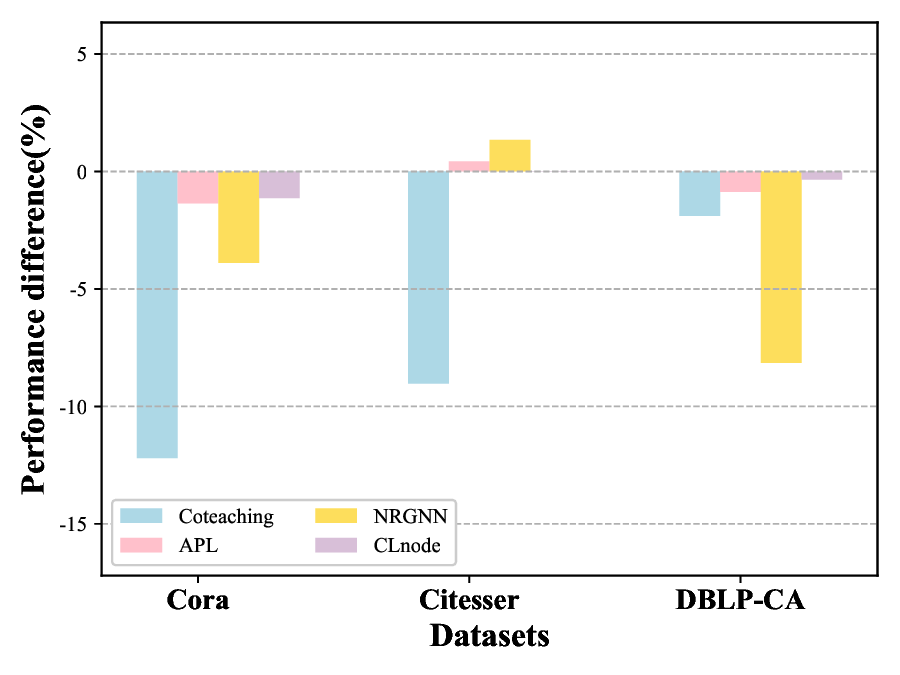}}
\subfigure[Random(0.3)]
{
\includegraphics[width=5.5cm,height =4.12cm]{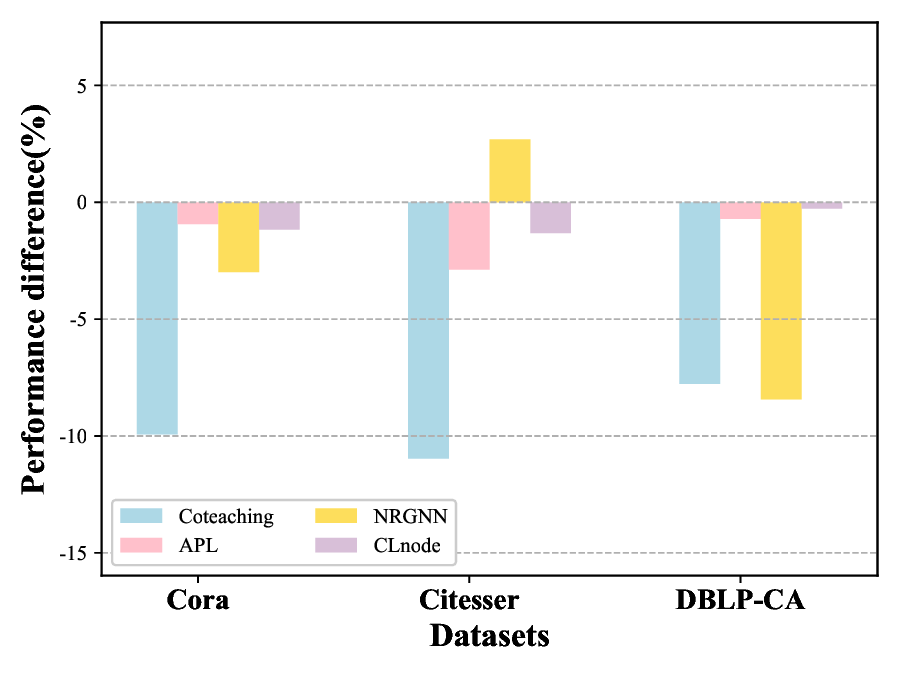}}
\caption{(a)–(c) illustrate the performance differences of different methods relative to HGNN across multiple datasets under 30\% pair, uniform and random noise, respectively.}
\label{fig:Preliminary}
\end{figure*}

\subsection{Empirical Results}
Under the proposed benchmark, we first evaluate several representative methods under a unified HGNN backbone and compare their performance against HGNN, as shown in Fig.~\ref{fig:Preliminary}. The x-axis represents different datasets and the y-axis represents the performance difference relative to HGNN. The missing purple bars in Fig.~\ref{fig:Preliminary}(a) indicate that the performance of CLNode and HGNN on the Citeseer dataset is nearly identical. Several important observations can be drawn from these results.
\begin{itemize}
    \item Existing methods yield only limited improvements over HGNN and frequently cause performance degradation, indicating that their robustness mechanisms are inadequate for hypergraph settings.
    \item Both LLN methods (e.g., Co-teaching\cite{Yu_Han_Yao_Niu_Tsang_Sugiyama_2019} and APL~\cite{ma2020normalized}) and GLN methods (e.g., NRGNN~\cite{dai2021nrgnn} and CLNode~\cite{wei2023clnode}) exhibit substantial variability across datasets, indicating limited cross-dataset consistency and weak generalization.
    \item Similar patterns are observed across all three noise types, suggesting that the limitation is rooted not in a specific noise pattern, but in the inadequate modeling for hypergraph structures.
\end{itemize}
Overall, these findings highlight the insufficiency of  LLN and GLN methods for robust hypergraph learning and  further motivate our study of robust learning for hypergraphs under label noise.

\begin{figure*}[t]
\centering  
\includegraphics[width=17cm,height =11.9cm]{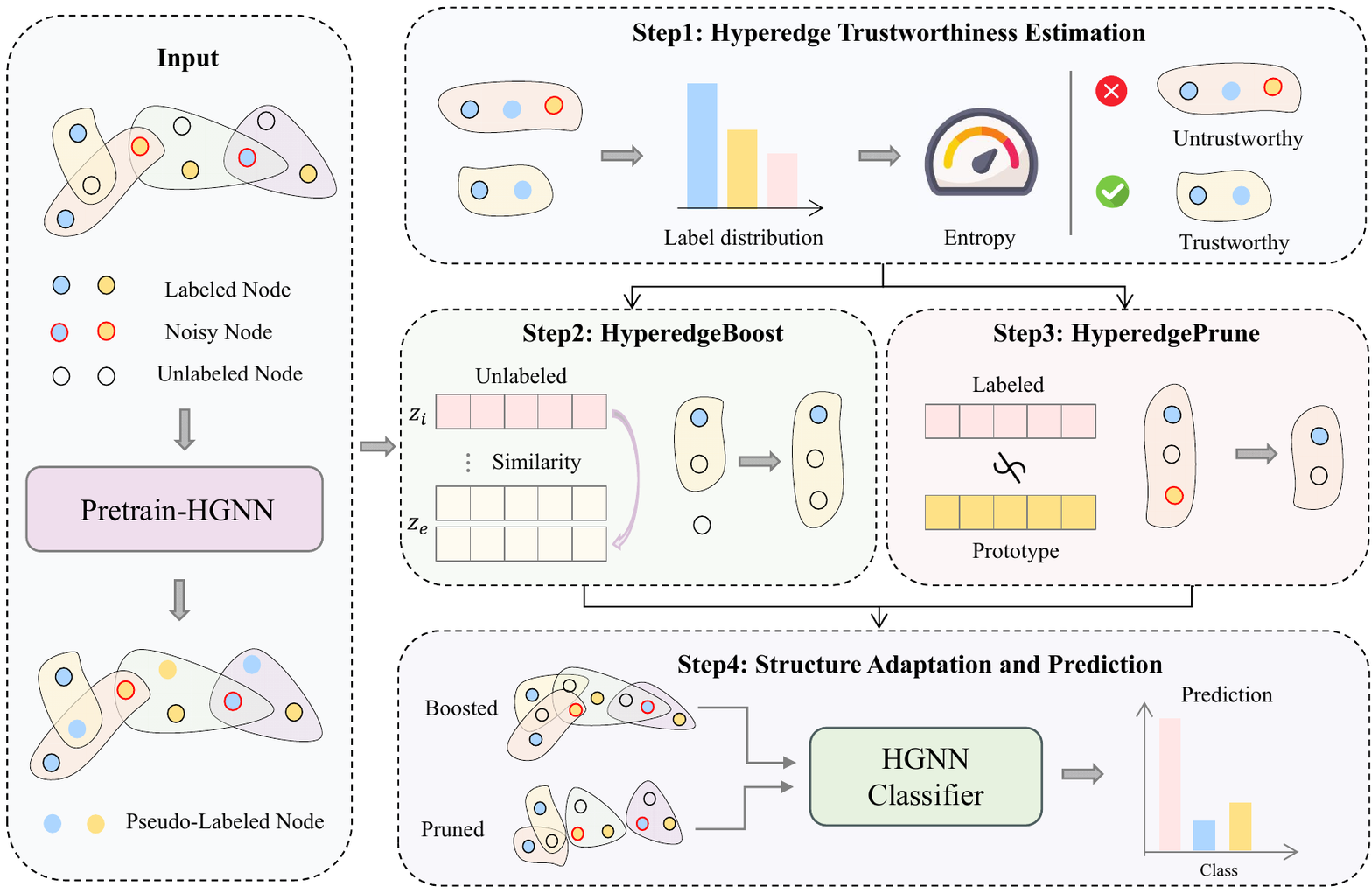}
 \caption{The framework of HyperTrust.}
\label{Framework}
\end{figure*}
\section{Methodology}
Based on the empirical observations in Section~\ref{Benchmark Design and Empirical Motivation}, and motivated by the unique high-order structure of hypergraphs, we aim to fully exploit hyperedge relationships by making  adjustments on the hypergraph. Specifically, we propose a novel model, named HyperTrust, to improve node classification on hypergraphs under label noise. HyperTrust consists of three main modules: Entropy-aware Hyperedge Trustworthiness Estimation, HyperedgeBoost, and HyperedgePrune. The overall framework of  HyperTrust is shown in Fig.~\ref{Framework}.

HyperTrust first pre-trains a hypergraph classifier on the hypergraph to obtain node embeddings and pseudo-labels for unlabeled nodes. It then estimates the trustworthiness of each hyperedge according to the label entropy of nodes within the hyperedge (see Section~\ref{Entropy-aware Hyperedge Trustworthiness Estimation}). Based on the estimated trustworthiness, HyperTrust performs HyperedgeBoost to connect unlabeled nodes to trustworthy hyperedges, thereby supplementing useful high-order message-passing paths (see Section~\ref{HyperedgeBoost Module}). Meanwhile, HyperTrust performs HyperedgePrune on untrustworthy hyperedges to remove noisy nodes associated with these hyperedges and block noisy message propagation (see Section~\ref{HyperedgePrune Module}). Finally, a shared hypergraph neural network makes predictions on both the boosted and pruned hypergraphs, and the two predictions are fused to generate the final output.

\subsection{Entropy-aware Hyperedge Trustworthiness Estimation}
\label{Entropy-aware Hyperedge Trustworthiness Estimation}
In hypergraph learning, hyperedges serve as fundamental structures for high-order message passing by connecting multiple nodes. Under the homophily assumption, hyperedges are more likely to connect nodes belonging to the same semantic category. Therefore, a hyperedge whose incident nodes exhibit consistent labels is less likely to contain noisy nodes and can be regarded as a trustworthy structure for information propagation. In contrast, if the nodes within a hyperedge have inconsistent labels, the hyperedge is likely to be contaminated by noisy nodes, which may introduce semantic conflicts and degrade the quality of message passing. Therefore, it is essential to assess the trustworthiness of hyperedges before performing topology adaptation, so that trustworthy structures can be emphasized while noisy ones are suppressed.

Since  labels are not fully available during training, we use pseudo-labels to approximate node categories. Specifically, we first pre-train a hypergraph classifier on the hypergraph to obtain node representations and pseudo-labels, which serve as the basis for hyperedge trustworthiness estimation. The formulation is as follows:
\begin{equation}
    Z = f_{\text{pre}}(X, H),
    \label{2}
\end{equation}
followed by a classifier that generates predicted logits:
\begin{equation}
     Y^{pre} = f_{\text{cls}}(Z).
\end{equation}
The pseudo-label of node $v_i$ is obtained by:
\begin{equation}
\hat{y}_i = \arg\max_{c} Y^{pre}_{i,c}.
\end{equation}
Here, $f_{\text{pre}}$ denotes the encoder, for which we adopt an HGNN, and $f_{\text{cls}}$ denotes the classifier, implemented as an MLP. $Z$ denotes the node embeddings, $Y^{pre}$ denotes the predicted logits, and $\hat{y}_i$ denotes the pseudo-label obtained via argmax. The model is trained on labeled nodes under label noise:
\begin{equation}
\label{5}
    \mathcal{L}_{pre}
    =
    \frac{1}{|\mathcal{V}_{L}|}\sum_{v_i\in\mathcal{V}_{L}}
   l(y_i^{pre}, y_i^N).
\end{equation}
where $y^{pre}_i$ represents the predicted
logits of node $v_i$, $y^{N}_{i}$ represents the noisy label of node $v_{i}$ and $l(\cdot)$ is the cross-entropy loss.

After training, we  construct mixed labels by using observed labels for labeled nodes and pseudo-labels for unlabeled nodes:
\begin{equation}
\label{6}
    \tilde{y}_i =
    \begin{cases}
        y_i^{N}, & v_i\in\mathcal{V}_{L},\\
        \hat{y}_i, & v_i\notin\mathcal{V}_{L}.
    \end{cases}
\end{equation}

Based on $\tilde{y}_i$, we estimate hyperedge trustworthiness according to label consistency. For each hyperedge $e$, we compute its  label distribution:
\begin{equation}
\label{7}
    p_{c}^{(e)}
    =
    \frac{1}{|e|}
    \sum_{v_i\in e}
    \mathbb{I}(\tilde{y}_i=c),
\end{equation}
and measure its consistency using entropy:
\begin{equation}
\label{8}
    E(e)
    =
    -\sum_{c=1}^{C}p_{c}^{(e)}\log(p_{c}^{(e)}).
\end{equation}
where $C$ is the number of classes. A lower entropy indicates that the nodes in the hyperedge share more consistent labels, implying higher trustworthiness. In contrast, a higher entropy suggests that the hyperedge contains mixed semantic information and is potentially noisy. We then partition hyperedges into trustworthy and untrustworthy sets:
\begin{equation}
\label{9}
    \mathcal{E}_{trust}
    =
    \{e\in\mathcal{E}\mid {E}(e)\leq \delta\}, 
\end{equation}
\begin{equation}
\label{10}
    \mathcal{E}_{untrust}
    = \{e\in\mathcal{E}\mid E(e)> \delta\}.
\end{equation}
where $\delta$ is a small positive constant. Trustworthy hyperedges are regarded as trustworthy high-order structures and will be utilized in the HyperedgeBoost module, while untrustworthy hyperedges are considered noisy and will be further refined by the HyperedgePrune module.

\subsection{HyperedgeBoost Module}
\label{HyperedgeBoost Module}
In real-world hypergraph data, nodes with similar semantic and structural patterns are  likely to share the same label~\cite{jin2021node,kong2013inferring,grover2016node2vec}. However, due to incomplete observations, missing records, and imperfect hypergraph construction, such latent relations are not always fully reflected in the observed hyperedge structure. Consequently, some nodes may be highly consistent with existing hyperedges in terms of semantics and structure, yet their node-hyperedge incidence relations remain unobserved. For example, in a co-authorship hypergraph, a researcher may belong to an existing research group based on similar topics and collaborations, but is not included due to missing publication records~\cite{yadati2020nhp}. In a biological network, a gene may be functionally related to a known pathway, while its association is not captured due to experimental limitations~\cite{zitnik2024current}. This structural incompleteness weakens high-order message passing and limits the model's ability to provide trustworthy supervision signals for unlabeled nodes from reliable hyperedges. Effectively identifying these potential relational connections can enhance the data's robustness to some extent.

Inspired by related research~\cite{dai2021nrgnn,zhou2025graph}, we observe that introducing additional node--hyperedge incidence relations between unlabeled nodes and trustworthy hyperedges can provide more accurate supervision signals. This design leverages the high label consistency of trustworthy hyperedges while avoiding the introduction of uncertain high-order structures. Specifically, we first compute the embedding of each trustworthy hyperedge by averaging the representations of its incident nodes:
\begin{equation}
    z_e
    =
    \frac{1}{|e|}
    \sum_{v_i\in e}z_i,
    \label{eq:hgg_hyperedge_embedding}
\end{equation}
where $z_i$ denotes the node representation obtained from the pre-trained encoder. For each unlabeled node $v_i\in\mathcal{V}_{U}$ and each trustworthy hyperedge $e\in\mathcal{E}_{trust}$, we calculate the cosine similarity between the node embedding and the hyperedge embedding as follows:
\begin{equation}
    s_{i,e}
    =
    \frac{z_i^{\top}z_e}{\|z_i\|\|z_e\|}.
    \label{eq:hgg_node_edge_similarity}
\end{equation}
where $\|z_i\|$ and $\|z_e\|$ denote the  norms of $z_i$ and $z_e$, respectively. Then, for each unlabeled node, we select the top-$K$ most similar trustworthy hyperedges and add the corresponding incidence relations:
\begin{equation}
    H^{Boost}_{i,e} =
    \begin{cases}
        1, & H_{i,e}=1,\\
        1, &H_{i,e}=0, v_i\in\mathcal{V}_{U},\ e\in TopK_{\mathcal{E}_{trust}}(v_i),\\
        0, & \text{otherwise}.
    \end{cases}
    \label{eq:hgg_boost_incidence}
\end{equation}
where $TopK_{\mathcal{E}_{trust}}(v_i)$ denotes the set of $K$ trustworthy hyperedges with the highest similarity to node $v_i$. Then, we use an HGNN classifier to make predictions on the boosted hypergraph:
\begin{equation}
    Y^{Boost}
    =
    f_{\theta}(X,H^{Boost}),
    \label{eq:hgg_boost_prediction}
\end{equation}
where $f_{\theta}$ denotes the HGNN classifier, $Y^{Boost}$ denotes the prediction matrix on the boosted hypergraph. This module enables unlabeled nodes to receive information from reliable hyperedges, thereby enhancing useful high-order message propagation and improving the robustness of node representations.
\subsection{HyperedgePrune Module}
\label{HyperedgePrune Module}
Although HyperedgeBoost introduces additional  propagation paths for unlabeled nodes by incorporating trustworthy incidence relations, it does not completely eliminate the risk of noise propagation inherited from the hypergraph. In particular, if noisy nodes remain involved in message passing, erroneous information may still propagate to their neighboring nodes through existing connections.

To address this issue, we further propose a HyperedgePrune module, which aims to suppress noise propagation by selectively removing unreliable node--hyperedge incidence relations. Instead of directly pruning nodes or globally removing their connections, we adopt a conservative relation-level pruning strategy to avoid damaging informative structural patterns.

Specifically, we first construct a class prototype for each class based on node representations and mixed labels:
\begin{equation}
    \mu_c
    =
    \frac{1}{|\mathcal{V}_c|}
    \sum_{v_i\in\mathcal{V}_c}z_i,
    \quad
    \mathcal{V}_c=\{v_i\mid \tilde{y}_i=c\}.
    \label{eq:hgg_class_proto}
\end{equation}
where $\mu_c$ denotes the prototype of class $c$, and $\mathcal{V}_c$ is the set of nodes assigned to class $c$.

Based on these prototypes, we measure the consistency between each node and its assigned class. A low similarity between a node and its corresponding class prototype typically indicates a potential label or semantic inconsistency, suggesting that the node may be noisy. The similarity is computed as:
\begin{equation}
    a_i
    =
    \frac{z_i^{\top}\mu_{\tilde{y}_i}}{\|z_i\|\|\mu_{\tilde{y}_i}\|}.
    \label{eq:hgg_proto_similarity}
\end{equation}

However, directly removing all incidence relations of such nodes may disrupt valid message propagation paths and degrade the structural integrity of the hypergraph. Therefore, we restrict pruning to high-risk scenarios. Specifically, for each incidence relation $(v_i, e)$, we perform pruning only when the node $v_i$ is labeled and the hyperedge $e$ is identified as untrustworthy. This design is motivated by the fact that  untrustworthy hyperedges are more likely to serve as channels for noise propagation.

Specifically, we first normalize the
prototype consistency scores within \(V_L\):
\begin{equation}
\bar{a}_i =
\frac{a_i-\min_{v_j\in V_L}a_j}
{\max_{v_j\in V_L}a_j-\min_{v_j\in V_L}a_j},
\end{equation}
 A smaller
\(\bar{a}_i\) indicates lower consistency with the assigned class prototype. Then, if the similarity score $\bar{a}_i$ falls below the threshold $\rho$, the node is regarded as inconsistent with its assigned class, and its connection to the untrustworthy hyperedge is removed. The pruned incidence matrix is defined as:
\begin{equation}
    H^{\text{Prune}}_{i,e}
    =
    \begin{cases}
    0, & \text{if } v_i \in \mathcal{V}_{L},\ e \in \mathcal{E}_{\text{untrust}},\ H_{i,e}=1\\& \text{and } \bar{a}_i < \rho, \\
    H_{i,e}, & \text{otherwise}.
    \end{cases}
    \label{eq:hgg_prune_incidence}
\end{equation}

In this way, pruning is concentrated on high-risk noise propagation channels, while reliable structures within trustworthy hyperedges are preserved, thereby avoiding excessive distortion of the overall topology. 
Finally, the classifier $f_{\theta}$ performs prediction on the pruned hypergraph:
\begin{equation}
    Y^{\text{Prune}}
    =
   f_{\theta}(X, H^{\text{Prune}}).
    \label{eq:hgg_prune_prediction}
\end{equation}

To further leverage the complementary properties of the boosted and pruned hypergraphs, we fuse their prediction logits:
\begin{equation}
    Y^{\text{Final}}
    =
    \frac{1}{2} (Y^{\text{Boost}}
    +
   Y^{\text{Prune}})
    \label{eq:hgg_fusion}
\end{equation}

The final training objective is defined as:
\begin{equation}
\label{21}
    \mathcal{L}
    =
    \frac{1}{|\mathcal{V}_{L}|}\sum_{v_i\in\mathcal{V}_{L}}
   l(y_i^{Final}, y_i^N).
\end{equation}
$y_{i}^{Final}$ represents the prediction of node $v_{i}$, $y^{N}_{i}$ represents the noisy label of node $v_{i}$   and $l(\cdot)$ is the cross entropy loss.
Overall, HyperedgeBoost and HyperedgePrune construct two complementary hypergraph views. The boosted view enhances reliable information propagation by introducing missing yet trustworthy incidence relations, while the pruned view mitigates noise propagation by removing unreliable node--hyperedge connections. By integrating predictions from both views, the model effectively balances information enrichment and noise suppression, thereby improving the robustness of hypergraph-based node classification. The pseudo-code of the HyperTrust is shown in Appendix \ref{appa}.

\section{Theoretical Analysis}
\label{sec:theory}
In this section, we provide a linearized surrogate analysis of
the effectiveness of HyperTrust. Specifically, we first introduce the trustworthy message propagation margin to quantify the gap between class-consistent and class-inconsistent propagation weights. We then show that this margin  characterizes the expected propagated classification margin under  label noise. Finally, we prove that HyperTrust improves this margin under some assumptions.

\subsection{Trustworthy Message Propagation Margin}
Consider a hypergraph $\mathcal{H}$ with $n$ nodes and $C$ classes.
Let $y_i$ and $y_i^N$ denote the true label and observed noisy label of node $v_i$, respectively. Under the linear propagation assumption,
let $P\in\mathbb{R}^{n\times n}$ and $\sum_j P_{ij}=1$ be the propagation matrix of hypergraph $\mathcal{H}$, where $P_{ij}$ measures the contribution of node $j$ to node $i$. Under label noise, the propagated  label message at node $v_i$ is defined as:
\begin{equation}
m_i(P)=\sum_{j=1}^n P_{ij} e_{ y_j^N}.
\end{equation}
where $e_{y_j^N} \in \{0, 1\}^C$ is the one-hot encoding of the observed noisy label $y_j^N$. To understand how noise affects propagation, we analyze the margin between the true class and incorrect classes after propagation.

For a node $v_i$ with true label $y_i=c$, consider the pairwise propagated classification margin between class $c$ and an incorrect class $r\neq c$:
\begin{equation}
m_i(P)_c - m_i(P)_r.
\end{equation}

We first analyze the propagation matrix $P$ by partitioning the propagation weights based on the nodes' true labels.
\begin{Definition}[Class-consistent and class-inconsistent propagation weights]
For a node $i$ with true label $y_i=c$, define
\begin{equation}
S_i^+(P)=\sum_{j:y_j=c}P_{ij},\qquad
S_{i,r}^-(P)=\sum_{j:y_j=r}P_{ij},\quad r\neq c.
\end{equation}
\end{Definition}

Based on this definition, the following theorem characterizes the expected classification margin under uniform label noise.

\begin{Theorem}
\label{thm:noise-margin}
For node $v_i$ with $y_i=c$, under uniform label noise with noise rate $\epsilon$, if
\begin{equation}
\lambda=1-\epsilon-\frac{\epsilon}{C-1}>0,
\end{equation}
then for any $r\neq c$, the expected   classification margin  satisfies
\begin{equation}
\mathbb{E}\big[m_i(P)_c-m_i(P)_r\mid Y\big]
=
\lambda \left(S_i^+(P)-S_{i,r}^-(P)\right).
\end{equation}
\end{Theorem}

The proof of Theorem~\ref{thm:noise-margin} is shown in  Appendix~\ref{appb}.For other types of label noise, the corresponding analysis follows a similar reasoning process with the appropriate label transition probabilities. Further details are provided in Appendix~\ref{appb}. Since correct classification requires the true class to dominate all incorrect classes, we consider the worst-case margin across all $r\neq c$.

\begin{Definition}[Trustworthy message propagation margin]
For a node $i$ with $y_i=c$, define
\begin{equation}
\Gamma_i(P)=S_i^+(P)-\max_{r\neq c}S_{i,r}^-(P).
\end{equation}
\end{Definition}

Combining Definition 2 with Theorem~\ref{thm:noise-margin}, we obtain
\begin{equation}
\min_{r\neq c}\mathbb{E}\big[m_i(P)_c-m_i(P)_r\mid Y\big]
=
\lambda \Gamma_i(P).
\end{equation}

This result shows that, when $\lambda>0$, increasing $\Gamma_i(P)$ directly enlarges the worst-case expected classification margin after noisy propagation.

\subsection{Robustness Analysis of HyperTrust}
For a hyperedge $e$, let $p^{(e)}=(p_1^{(e)},\dots,p_C^{(e)})$ denote its class proportion vector. The entropy of 
$e$ is defined as described in Section~\ref{Entropy-aware Hyperedge Trustworthiness Estimation}. Let $q_e=\max_c p_c^{(e)}$ represent the purity of $e$. A lower entropy indicates a more concentrated class distribution, implying higher purity and stronger semantic consistency. Based on this observation, we analyze how HyperTrust improves propagation robustness.

Under the assumption that low-entropy hyperedges are less likely to contain noisy-labeled nodes,  the HyperedgeBoost module connects nodes to trustworthy low-entropy hyperedges. As a result, the additional propagation paths introduced by HyperedgeBoost are more likely to be class-consistent, which improves the trustworthy message propagation margin in expectation.

\begin{Lemma} 
\label{lem:add}
For node $v_i$ with $y_i=c$, suppose that the HyperedgeBoost module injects a propagation weight $\eta_i\in(0,1]$ from trustworthy hyperedges whose purity is at least $q_\star$ and whose dominant class is $c$. 
Furthermore, assume that
\begin{equation}
2q_\star-1>\Gamma_i(P),
\end{equation}
Then, we have
\begin{equation}
\Gamma_i(P^{\mathrm{Boost}})
\ge
(1-\eta_i)\Gamma_i(P)+\eta_i(2q_\star-1)>\Gamma_i(P).
\end{equation}
\end{Lemma}

Lemma~\ref{lem:add} shows that HyperedgeBoost improves the trustworthy message propagation margin when the injected trustworthy hyperedges are sufficiently pure and aligned with the true class. The proof is provided in Appendix~\ref{appc}.

The HyperedgePrune module removes untrustworthy node--hyperedge relations according to the similarity between node representations and class prototypes. Since the pruning process does not access the true labels, its effectiveness depends on whether prototype-guided pruning can preferentially suppress class-inconsistent propagation weights. We characterize this effect with the following conditional result.

\begin{Lemma}
\label{thm:drop}
For node $v_i$ with $y_i=c$, suppose the HyperedgePrune module removes a total propagation weight $\omega_i\in[0,1)$.
Let $\alpha_i$ denote the removed propagation weight from class-consistent nodes, and let $\beta_{i,r}$ denote the removed propagation weight from nodes of each incorrect class $r\neq c$. Assume that the strongest competing class remains unchanged after pruning, i.e.,
\begin{equation}
r^\star
=
\arg\max_{r\neq c} S_{i,r}^-(P)
=
\arg\max_{r\neq c} S_{i,r}^-(P^{\mathrm{Prune}}).
\end{equation}
Furthermore, assume that the original margin is non-negative: $\Gamma_i(P)\geq 0$.
If the prototype-guided pruning is class-selective in the sense that it removes more propagation weight from the strongest competing class than from the true class, i.e.,
\begin{equation}
\beta_{i,r^\star}>\alpha_i,
\end{equation}
then, after renormalization, the trustworthy message propagation margin is strictly improved:
\begin{equation}
\Gamma_i(P^{\mathrm{Prune}})>\Gamma_i(P).
\end{equation}
\end{Lemma}

Lemma~\ref{thm:drop} shows that HyperedgePrune improves the trustworthy message propagation margin when the prototype-based pruning criterion is effective in identifying unreliable node--hyperedge relations. This condition does not require the pruning module to access the true labels; rather, it characterizes the desired effect that nodes inconsistent with their corresponding class prototypes are more likely to be removed, so that the removed propagation mass concentrates more on class-inconsistent nodes than on class-consistent nodes. The proof is provided in Appendix~\ref{appc}.

Finally, HyperTrust combines the predictions from the boosted and
pruned views through a fusion:
\begin{equation} Y^{\mathrm{Final}} = \frac{1}{2} \left( Y^{\mathrm{Boost}} + Y^{\mathrm{Prune}} \right). \label{eq:prediction-fusion} \end{equation}
To analyze this fusion under the linear propagation setting, we
define the corresponding surrogate propagation operator as
\begin{equation}
P^{\mathrm{Fuse}}
=
\frac{1}{2}
\left(
P^{\mathrm{Boost}}
+
P^{\mathrm{Prune}}
\right).
\label{eq:propagation-fusion}
\end{equation}

Since the propagated label message $m_i(P)$ is linear with
respect to $P$, we have
\begin{equation}
m_i(P^{\mathrm{Fuse}})
=
\frac{1}{2}
\left[
m_i(P^{\mathrm{Boost}})
+
m_i(P^{\mathrm{Prune}})
\right].
\label{eq:message-fusion}
\end{equation}

The following theorem shows that the fused surrogate operator
preserves the margin improvements from the two views.

\begin{Theorem}
\label{prop:fusion}
For any node $v_i$, under the assumptions of
Lemma~\ref{lem:add} and Lemma~\ref{thm:drop}, we have
\begin{equation}
\Gamma_i(P^{\mathrm{Fuse}})
\geq
\frac{1}{2}
\left[
\Gamma_i(P^{\mathrm{Boost}})
+
\Gamma_i(P^{\mathrm{Prune}})
\right]
>
\Gamma_i(P).
\end{equation}
\end{Theorem}

The proof of Theorem~\ref{prop:fusion} is provided in
Appendix~\ref{appc}. Theorem~\ref{prop:fusion} shows that the
convex fusion preserves the complementary benefits of
HyperedgeBoost and HyperedgePrune under the linear propagation
analysis. Here, $P^{\mathrm{Fuse}}$ is introduced only as a
surrogate operator for theoretical analysis, while the actual
model performs prediction-level fusion as defined in
Eq.~\eqref{eq:prediction-fusion}.

\section{Experiments}
\subsection{Experiment Setup}
\subsubsection{Dataset}
\begin{table*}[htbp]
  \centering
  \caption{Dataset Statistics Summary\label{tab:dataset_stats}}
  \normalsize
  \setlength{\tabcolsep}{4pt} 
  \begin{tabular}{l *{7}{S[table-format=5.0]}} 
    \toprule
    {\textbf{Metric}} & {\textbf{Cora}} & {\textbf{Citeseer}} & {\textbf{Pubmed}} & {\textbf{Cora-CA}} & {\textbf{DBLP-CA}}   & {\textbf{NTU2012}} & {\textbf{ModelNet40}}  \\
    \midrule
    $|V|$ & 2708 & 3312 & 19177 & 2708 & 41302  & 2012 & 12311   \\
    $|E|$ & 1579 & 1079 & 7963 & 1072 & 22363  &  2012 & 12311  \\
    \#features
    & 1433 & 3703 & 500 & 1433 & 1425  & 100  & 100   \\
    \#classes & 7 & 6 & 3  & 7 & 6  &  67 & 40 \\
    \bottomrule
  \end{tabular}
  \vspace{0.2cm}
\end{table*}
To comprehensively evaluate the performance of HyperTrust, LLN, and GLN, we conduct extensive experiments on a diverse collection of benchmark datasets spanning both academic and real-world scenarios. For academical scenarios, we use five commonly
adopted hypergraph benchmarks constructed from co-citation and
co-authorship networks: Cora, Citeseer, Pubmed, Cora-CA, and
DBLP-CA~\cite{yadati2019hypergcn}~\cite{zhou2026tackling}. In these datasets, node features are
given by bag-of-words representations, while labels denote the subject
categories of papers.

To investigate the generalizability of different models beyond academic networks,
we further include two real-world datasets from different domains,
including  the 3D vision
datasets ModelNet40~\cite{wu20153d} and NTU2012~\cite{chen2003visual}. For fair comparison, all methods are evaluated on the same hypergraph structure constructed following the protocol in prior work.
Each dataset is split into training, validation, and test sets with a
50\%/25\%/25\% ratio. All experiments are repeated ten times, each with independently generated noisy labels, and the reported results are the mean performance across these ten runs. The statistics of all datasets are provided in Table~\ref{tab:dataset_stats}.

\subsubsection{Label corruption}
We follow the label-noise settings adopted in~\cite{wang2024noisygl,zhou2025graph}.
For each experiment, we first construct a label transition probability
matrix according to the specified noise rate and the corresponding noise
generation protocol. Each row of this matrix characterizes the
class-conditional distribution of the observed noisy label given the
underlying clean label. Then, for every sample in the training and
validation sets, we generate its noisy label by sampling from the categorical distribution specified by the corresponding row of the transition matrix. The generated
noisy labels are fixed throughout the experiment and are used as the
supervision signals during the entire training process.

\subsubsection{Implementations}

We report the average accuracy and standard deviation over ten repeated experimental runs. Both the pre-training classifier and the final classifier employ HGNN as the backbone network. 
In all experiments,  $\delta$ is set to $1e-10$, a very small value chosen to ensure that only highly trustworthy hyperedges are selected. For hyper-parameter tuning,  we search the number of nearest hyperedge candidates $K$ over the set
$\{5,10,20,30,40,50,60\}$ and the pruning threshold $\rho$ over 
$\{0.3,0.4,0.5,0.6,0.7,0.8,0.9\}$. The learning rate and weight decay are selected from the set
$\{1\mathrm{e}{-}1,5\mathrm{e}{-}2,1\mathrm{e}{-}2,5\mathrm{e}{-}3,
1\mathrm{e}{-}3,5\mathrm{e}{-}4\}$
and the set $\{5\mathrm{e}{-}2,5\mathrm{e}{-}3,5\mathrm{e}{-}4,5\mathrm{e}{-}5\}$, respectively. 
The number of layers is searched over $\{2,3,4,5\}$, while the hidden dimension is selected from 
$\{16,32,64,128,256\}$. Source code is available at \href{https://anonymous.4open.science/r/NoisyHGL-871D}{https://anonymous.4open.science/r/NoisyHGL-871D}.

\subsection{Performance comparison}
\subsubsection{Main results}
To comprehensively evaluate the performance of HyperTrust on hypergraph datasets, we adapt several representative robustness methods, such as LLN and GLN, to the hypergraph setting. All compared methods are implemented and evaluated under the same experimental protocol to ensure a fair comparison. Specifically, the following methods are considered as baselines for comparison with HyperTrust.

    \setlength{\parindent}{2em} $\bullet$ \textbf{HGNN~\cite{feng2019hypergraph}} is a type of Hypergraph Neural Network (HGNN) that perform spectral convolution on Hypergraph data.

\setlength{\parindent}{2em} $\bullet$ \textbf{ S-model ~\cite{goldberger2017training}} models label noise by adding a learnable noise adaptation layer during training, which is removed at test time to enable prediction on true labels.

    \setlength{\parindent}{2em} $\bullet$ \textbf{Co-teaching\cite{Yu_Han_Yao_Niu_Tsang_Sugiyama_2019}} trains two deep neural networks simultaneously, where each selects small-loss samples and passes them to the other.
    
    \setlength{\parindent}{2em} $\bullet$ \textbf{JoCoR\cite{wei2020combating}} addresses noisy labels by using a pseudo-twin network with two classifiers that are jointly trained to minimize prediction differences through a loss function combining supervised learning and contrast loss.

     \setlength{\parindent}{2em} $\bullet$ \textbf{APL~\cite{ma2020normalized}} constructs a robust objective by jointly combining active and passive loss terms, which helps the model maintain sufficient fitting ability while reducing the influence of noisy labels.

    \setlength{\parindent}{2em} $\bullet$ \textbf{Forward correction~\cite{Patrini2017over}} improves model robustness in noisy labeling environments by linearly combining the DNN's softmax output with the estimated label transfer probability before applying the loss function.

\setlength{\parindent}{2em} $\bullet$ \textbf{Backward correction~\cite{Patrini2017over}} mitigates label noise by estimating a label transition matrix and using it to correct the loss during model retraining.

   \setlength{\parindent}{2em} $\bullet$ \textbf{NRGNN~\cite{dai2021nrgnn}}  connecting nodes with similar features to create a refined graph. This refined graph is then used to generate precise pseudo-labels, enabling unlabeled nodes to receive more supervision from correctly labeled nodes and mitigating the impact of noisy labels.
   
    \setlength{\parindent}{2em} $\bullet$ \textbf{CP~\cite{zhang2020adversarial}} is a defense framework proposed to counteract adversarial label-flipping attacks on GCNs, using community labels as high-level signals to guide node classification.

    \setlength{\parindent}{2em} $\bullet$ \textbf{CLNode~\cite{wei2023clnode}} mitigates the impact of label noise by adopting a curriculum learning strategy .
    
    \setlength{\parindent}{2em} $\bullet$  \textbf{PIGNN\cite{du2021noise}}  uses a PI label estimation method based on node embeddings to constrain the training process.
    
    \setlength{\parindent}{2em} $\bullet$ \textbf{CGNN~\cite{yuan2023learning}} addresses label noise in GNNs by combining neighborhood based label correction with contrastive learning.

\begin{figure*}[h!]
\centering  
\subfigure[\textbf{Pair-AUCS}]
{
\includegraphics[width=5.5cm,height =5.2cm]{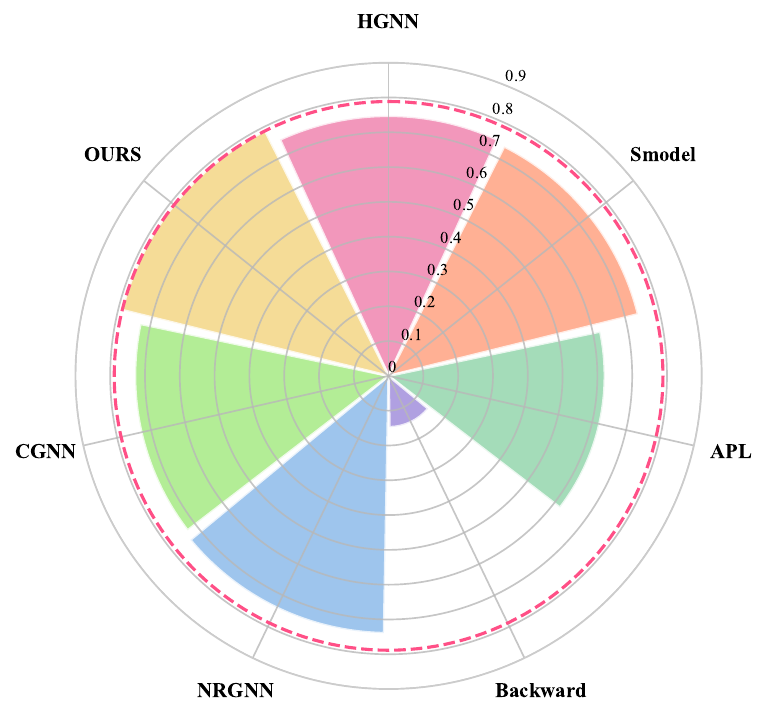}}
\subfigure[\textbf{Pair-AUU}]
{
\includegraphics[width=5.5cm,height =5.2cm]{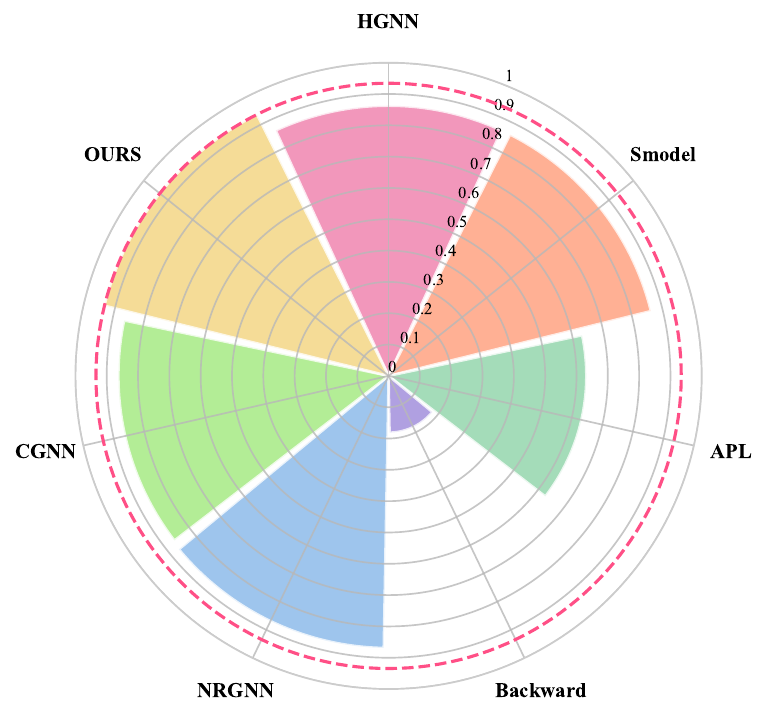}}
\subfigure[\textbf{Pair-AUIS}]
{
\includegraphics[width=5.5cm,height =5.2cm]{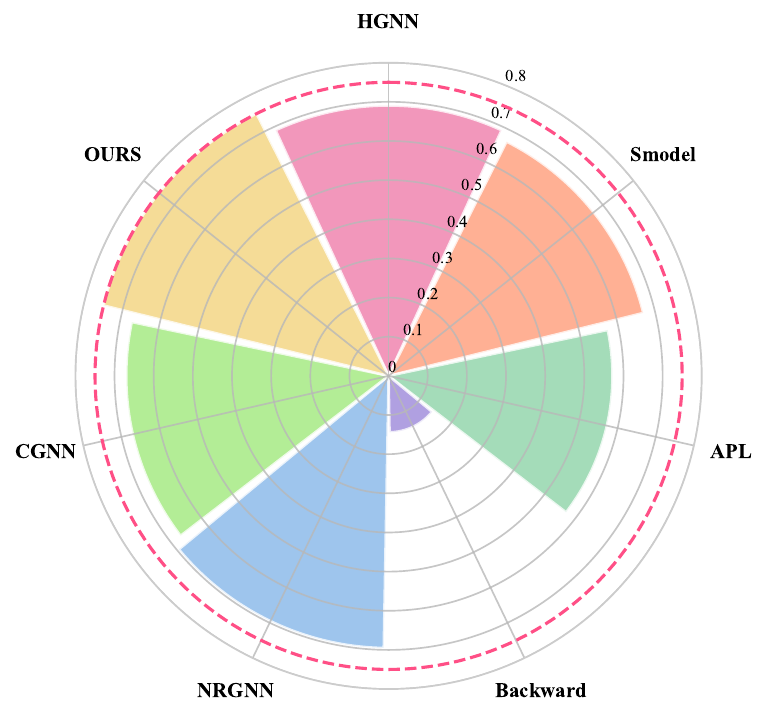}}
\subfigure[\textbf{Uniform-AUCS}]
{
\includegraphics[width=5.5cm,height =5.2cm]{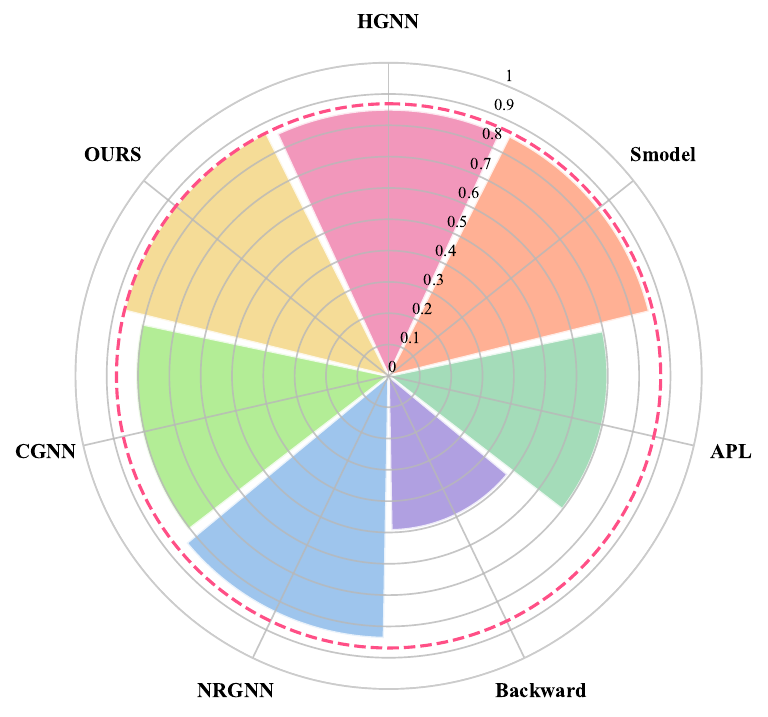}}
\subfigure[\textbf{Uniform-AUU}]
{
\includegraphics[width=5.5cm,height =5.2cm]{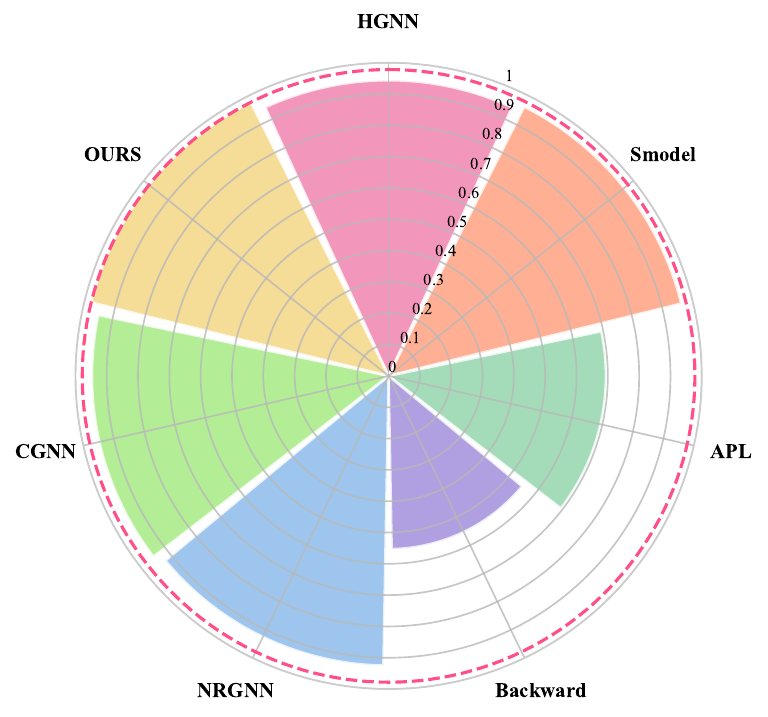}}
\subfigure[\textbf{Uniform-AUIS}]
{
\includegraphics[width=5.5cm,height =5.2cm]{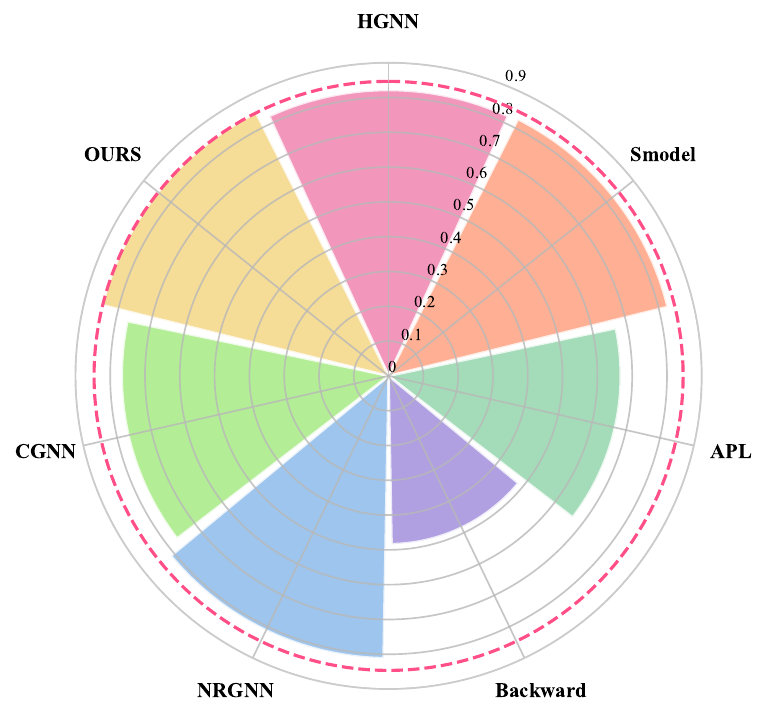}}
\caption{(a)(b)(c) show multi-metric results on the NTU2012 dataset under pair noise. (d)(e)(f) show multi-metric results on the NTU2012 dataset under uniform noise.}
\label{fig:MR}
\end{figure*}

\begin{table*}[t]
\centering
\caption{Node classification performance (Accuracy (\%)$\pm$Std) under various types of noise (rate 0.3).}
\label{tab:1}
\begin{adjustbox}{max width=\textwidth}
\begin{tabular}{c|ccc|ccc|ccc}
  \toprule
  \multirow{2}{*}{\centering Dataset} & \multicolumn{3}{c|}{Cora} & \multicolumn{3}{c|}{Citeseer} & \multicolumn{3}{c}{Pubmed} \\
   & pair & uniform & random & pair & uniform & random & pair & uniform & random \\
  \hline
  HGNN & \textcolor{OliveGreen}{\textbf{67.52 $\pm$ 3.34}} & \textcolor{OliveGreen}{\textbf{72.66 $\pm$ 1.53}} & \textcolor{OliveGreen}{\textbf{72.61 $\pm$ 1.71}} & \textcolor{OliveGreen}{\textbf{64.49 $\pm$ 2.03}} & 70.22 $\pm$ 1.61 & 68.54 $\pm$ 2.11 & 78.01 $\pm$ 0.76 & 82.51 $\pm$ 0.61 & 81.18 $\pm$ 1.42 \\
  \hline
  S-model & \textcolor{OliveGreen}{\textbf{67.42 $\pm$ 3.51}} & \textcolor{OliveGreen}{\textbf{73.35 $\pm$ 1.43}} & \textcolor{OliveGreen}{\textbf{72.96 $\pm$ 1.57}} & 63.94 $\pm$ 2.31 & 70.16 $\pm$ 1.40 & \textcolor{OliveGreen}{\textbf{68.95 $\pm$ 2.09}} & 78.99 $\pm$ 0.67 & 82.48 $\pm$ 0.56 & 82.02 $\pm$ 1.26 \\
  Co-teaching & 55.12 $\pm$ 12.05 & 60.46 $\pm$ 12.08 & 62.68 $\pm$ 6.18 & 54.34 $\pm$ 6.74 & 61.19 $\pm$ 7.15 & 57.57 $\pm$ 8.34 & 74.33 $\pm$ 8.37 & 81.31 $\pm$ 2.77 & 76.92 $\pm$ 8.59 \\
  JoCoR & 64.53 $\pm$ 5.96 & 59.83 $\pm$ 13.43 & 65.61 $\pm$ 8.62 & 58.81 $\pm$ 6.64 & 66.70 $\pm$ 7.68 & 63.33 $\pm$ 7.71 & \textcolor{OliveGreen}{\textbf{80.52 $\pm$ 0.86}} & 82.32 $\pm$ 0.52 & \textcolor{red}{\textbf{83.25 $\pm$ 1.40}} \\
  APL & 66.85 $\pm$ 5.27 & 71.30 $\pm$ 3.82 & 71.67 $\pm$ 2.71 & 61.64 $\pm$ 6.84 & 70.65 $\pm$ 3.11 & 65.66 $\pm$ 5.17 & 80.21 $\pm$ 0.59 & 82.49 $\pm$ 0.54 & \textcolor{OliveGreen}{\textbf{82.73 $\pm$ 0.47}} \\
  Forward & 62.80 $\pm$ 8.73 & 71.60 $\pm$ 3.03 & 65.71 $\pm$ 8.90 & 57.65 $\pm$ 8.82 & \textcolor{OliveGreen}{\textbf{71.04 $\pm$ 1.69}} & 67.25 $\pm$ 3.91 & 76.69 $\pm$ 9.34 & 82.34 $\pm$ 0.85 & 81.90 $\pm$ 1.38 \\
  \hline
  NRGNN & 63.94 $\pm$ 2.41 & 68.77 $\pm$ 1.30 & 69.62 $\pm$ 1.04 & \textcolor{OliveGreen}{\textbf{66.28 $\pm$ 1.61}} & \textcolor{OliveGreen}{\textbf{71.56 $\pm$ 1.07}} & \textcolor{OliveGreen}{\textbf{71.12  $\pm$ 1.78}} & 77.40 $\pm$ 0.37 & 80.65 $\pm$ 0.46 & 79.97 $\pm$ 1.73 \\
  CP & 60.68 $\pm$ 3.69 & 68.64 $\pm$ 3.76 & 68.41 $\pm$ 3.88 & 55.96 $\pm$ 4.28 & 63.39 $\pm$ 3.57 & 60.42 $\pm$ 4.05 & 77.78 $\pm$ 1.06 & 82.28 $\pm$ 0.48 & 81.03 $\pm$ 1.30 \\
  CLNode & 65.51 $\pm$ 3.16 & 71.52 $\pm$ 1.84 & 71.44 $\pm$ 2.82 & 63.25 $\pm$ 3.63 & 70.23 $\pm$ 1.84 & 67.22 $\pm$ 2.68 & 79.77 $\pm$ 0.95 & \textcolor{OliveGreen}{\textbf{82.55 $\pm$ 0.43}} & 81.94 $\pm$ 1.65 \\
  PIGNN & 64.46 $\pm$ 4.36 & 68.23 $\pm$ 2.52 & 67.49 $\pm$ 2.65 & 61.63 $\pm$ 2.58 & 66.89 $\pm$ 2.43 & 66.81 $\pm$ 1.57 & \textcolor{OliveGreen}{\textbf{80.48 $\pm$ 1.38}} & \textcolor{OliveGreen}{\textbf{82.59 $\pm$ 0.59}} & 82.59 $\pm$ 1.12 \\
  CGNN & 60.81 $\pm$ 3.77 & 66.32 $\pm$ 4.67 & 66.71 $\pm$ 3.58 & 31.57 $\pm$ 10.10 & 32.80 $\pm$ 7.76 & 36.30 $\pm$ 13.16 & 57.95 $\pm$ 24.73 & 57.12 $\pm$ 25.26 & 56.99 $\pm$ 24.16 \\
  OURS & \textcolor{red}{\textbf{70.43 $\pm$ 2.16$^*$}} & \textcolor{red}{\textbf{74.75 $\pm$ 0.98$^*$}} & \textcolor{red}{\textbf{75.06 $\pm$ 1.95$^*$}} & \textcolor{red}{\textbf{68.54 $\pm$ 2.67$^*$}} & \textcolor{red}{\textbf{73.18 $\pm$ 0.90$^*$}} & \textcolor{red}{\textbf{72.18 $\pm$ 2.03}} & \textcolor{red}{\textbf{81.54 $\pm$ 1.20$^*$}} & \textcolor{red}{\textbf{83.72 $\pm$ 0.75$^*$}} & \textcolor{OliveGreen}{\textbf{82.88 $\pm$ 1.72}} \\
  \bottomrule
\end{tabular}
\end{adjustbox}
 \captionsetup{labelformat=empty}
\caption*{\scriptsize (a) Results on  Cora, Citeseer, and Pubmed.}

 \begin{adjustbox}{max width=\textwidth}
 \begin{tabular}{c|ccc|ccc|ccc}
  \toprule
  \multirow{2}{*}{\centering Dataset} & \multicolumn{3}{c|}{Cora-CA} & \multicolumn{3}{c|}{DBLP-CA} & \multicolumn{3}{c}{ModelNet40} \\
   & pair & uniform & random & pair & uniform & random & pair & uniform & random \\
  \hline
  HGNN & 71.83 $\pm$ 2.15 & 76.81 $\pm$ 1.54 & 76.65 $\pm$ 1.31 & 85.29 $\pm$ 0.29 & \textcolor{OliveGreen}{\textbf{89.59 $\pm$ 0.19}} & \textcolor{OliveGreen}{\textbf{89.33 $\pm$ 0.25}} & 96.44 $\pm$ 0.93 & 98.31 $\pm$ 0.16 & 98.43 $\pm$ 0.14 \\
  \hline
  S-model & 71.95 $\pm$ 2.19 & 76.46 $\pm$ 1.89 & \textcolor{OliveGreen}{\textbf{76.80 $\pm$ 1.25}} & \textcolor{OliveGreen}{\textbf{85.50 $\pm$ 0.21}} & \textcolor{OliveGreen}{\textbf{89.50 $\pm$ 0.17}} & \textcolor{OliveGreen}{\textbf{89.19 $\pm$ 0.21}} & 96.56 $\pm$ 1.01 & 98.33 $\pm$ 0.19 & 98.32 $\pm$ 0.15 \\
  Co-teaching & 60.81 $\pm$ 12.05 & 65.41 $\pm$ 12.86 & 67.63 $\pm$ 10.51 & 74.16 $\pm$ 13.35 & 87.61 $\pm$ 1.06 & 81.56 $\pm$ 10.83 & 90.71 $\pm$ 5.38 & 98.13 $\pm$ 0.19 & 98.09 $\pm$ 0.38 \\
  JoCoR & 67.52 $\pm$ 11.15 & 72.02 $\pm$ 10.88 & 66.77 $\pm$ 8.72 & 78.92 $\pm$ 10.56 & 83.36 $\pm$ 10.46 & 74.46 $\pm$ 12.03 & 88.56 $\pm$ 7.24 & 98.18 $\pm$ 0.15 & 98.11 $\pm$ 0.13 \\
  APL & \textcolor{OliveGreen}{\textbf{72.21 $\pm$ 4.09}} & 76.68 $\pm$ 1.84 & 76.48 $\pm$ 2.91 & \textcolor{OliveGreen}{\textbf{86.12 $\pm$ 0.20}} & 88.72 $\pm$ 0.20 & 88.62 $\pm$ 0.22 & 96.16 $\pm$ 3.45 & 98.36 $\pm$ 0.18 & 98.26 $\pm$ 0.27 \\
  Forward & \textcolor{OliveGreen}{\textbf{72.90 $\pm$ 2.02}} & \textcolor{OliveGreen}{\textbf{77.01 $\pm$ 1.47}} & 75.54 $\pm$ 2.02 & 79.06 $\pm$ 7.28 & 87.37 $\pm$ 0.72 & 86.88 $\pm$ 0.80 & 95.11 $\pm$ 3.44 & 98.06 $\pm$ 1.01 & 98.49 $\pm$ 0.41 \\
  \hline
  NRGNN & 64.86 $\pm$ 2.22 & 68.01 $\pm$ 2.30 & 68.67 $\pm$ 1.30 & 75.46 $\pm$ 0.76 & 81.44 $\pm$ 0.35 & 80.90 $\pm$ 0.93 & 96.19 $\pm$ 0.55 & 96.88 $\pm$ 0.13 & 96.76 $\pm$ 0.22 \\
  CP & 65.31 $\pm$ 3.15 & 75.06 $\pm$ 1.90 & 75.28 $\pm$ 1.79 & 83.68 $\pm$ 0.88 & 89.22 $\pm$ 0.30 & 88.90 $\pm$ 0.40 & 94.73 $\pm$ 0.96 & 98.32 $\pm$ 0.14 & 98.37 $\pm$ 0.17 \\
  CLNode & 67.62 $\pm$ 9.63 & 76.54 $\pm$ 1.69 & 76.20 $\pm$ 1.56 & 85.12 $\pm$ 0.82 & 89.24 $\pm$ 0.12 & 89.06 $\pm$ 0.47 & \textcolor{OliveGreen}{\textbf{97.36 $\pm$ 0.44}} & \textcolor{OliveGreen}{\textbf{98.51 $\pm$ 0.05}} & \textcolor{OliveGreen}{\textbf{98.55 $\pm$ 0.19}} \\
  PIGNN & 72.12 $\pm$ 2.35 & \textcolor{OliveGreen}{\textbf{77.09 $\pm$ 1.38}} & \textcolor{OliveGreen}{\textbf{76.72 $\pm$ 1.85}} & - & - & - & \textcolor{OliveGreen}{\textbf{97.81 $\pm$ 0.32}} & \textcolor{OliveGreen}{\textbf{98.63 $\pm$ 0.15}} & \textcolor{OliveGreen}{\textbf{98.60 $\pm$ 0.17}} \\
  CGNN & 39.61 $\pm$ 26.26 & 32.51 $\pm$ 23.75 & 28.10 $\pm$ 19.25 & 36.67 $\pm$ 11.20 & 39.56 $\pm$ 15.52 & 36.54 $\pm$ 16.98 & 77.97 $\pm$ 17.04 & 87.60 $\pm$ 14.65 & 86.96 $\pm$ 15.08 \\
  OURS & \textcolor{red}{\textbf{73.74 $\pm$ 1.79}} & \textcolor{red}{\textbf{79.64 $\pm$ 1.61$^*$}} & \textcolor{red}{\textbf{78.80 $\pm$ 1.95$^*$}} & \textcolor{red}{\textbf{88.01  $\pm$ 0.31$^*$}} & \textcolor{red}{\textbf{89.96 $\pm$ 0.26$^*$}} & \textcolor{red}{\textbf{89.73 $\pm$ 0.32$^*$}} & \textcolor{red}{\textbf{98.33 $\pm$ 0.21$^*$}} & \textcolor{red}{\textbf{99.08 $\pm$ 0.09$^*$}} & \textcolor{red}{\textbf{98.99 $\pm$ 0.20$^*$}} \\
  \bottomrule
\end{tabular}

\end{adjustbox}
\vspace{0.2cm}
\caption*{\scriptsize (b) Results on Cora-CA, DBLP-CA, and ModelNet40.}
\end{table*}
To ensure a fair and rigorous comparison, all baseline results were obtained using a unified dataset partition and a standardized HGNN backbone. Each experiment was repeated over ten runs under identical configurations but with different random seeds, where the seed determines the noisy-label generation. In this section, we present a representative subset of our extensive evaluation, specifically reporting the performance under pair, uniform, and random noise at a fixed noise rate of 0.3, as summarized in Table \ref{tab:1}. In this table, the optimal results are highlighted in red while the second and third best performances are indicated in green.  Furthermore, an asterisk ($^*$) denotes that the improvement of HyperTrust over the second-best baseline is statistically significant ($p < 0.05$) based on a two-sample t-test. Due to space constraints, comprehensive experimental results including a broader range of noise rates, additional datasets, and further comparisons with baseline methods are provided in the source code. From the results presented in Table \ref{tab:1}, several key observations can be made:

 \begin{itemize}
    \item Existing LLN and GLN methods, including Co-teaching~\cite{Yu_Han_Yao_Niu_Tsang_Sugiyama_2019}, NRGNN~\cite{dai2021nrgnn}, and CLNode~\cite{wei2023clnode}, show suboptimal performance on hypergraph datasets, indicating their limited ability to handle label noise in hypergraph learning. The results also show that pair noise causes the largest performance degradation across all benchmarks, as it introduces class-conditional interference that is harder to distinguish from true label patterns than uniform or random noise.

    \item HyperTrust outperforms competing methods on most datasets, showing strong robustness on large-scale hypergraphs such as DBLP-CA and ModelNet40. This advantage can be attributed to its ability to leverage high-order relations for reliable supervision, which improves structural adaptability and helps maintain high classification accuracy even when baseline models degrade substantially.

    \item HyperTrust achieves the best or competitive performance in most settings, demonstrating exceptional resilience, particularly under the challenging pair noise setting. This consistent advantage shows that, unlike existing methods that are often tailored to specific noise patterns, HyperTrust generalizes well across different corruption scenarios and offers a stable and effective framework for noisy hypergraph learning.

\end{itemize}
To provide a more fine-grained evaluation of HyperTrust, we further analyzed the NTU2012 dataset under 0.3  noise using several auxiliary metrics, including the accuracy of correctly labeled (ACLT), incorrectly labeled (AILT), and misleadingly labeled (AILMT) training nodes, as well as the performance on unlabeled nodes categorized by their local supervision quality. Specifically, we evaluate the accuracy of unlabeled correctly supervised (AUCS), unsupervised (AUU), and incorrectly supervised (AUIS) nodes, where correctly supervised, incorrectly supervised, and unsupervised refer to unlabeled samples that have at least one correctly labeled training node, an incorrectly labeled training node, or no labeled training nodes within their neighborhood, respectively. The results, summarized in Table \ref{tab:mulperformance} and Fig. \ref{fig:MR}, show that HyperTrust outperforms competing methods on almost all metrics. Although Backward achieves a low AILMT under pair noise, this result should be interpreted with caution, as its performance on other metrics is substantially lower, suggesting possible model collapse. In contrast, HyperTrust  performs well across fine-grained metrics. Its high AILT and AUIS scores indicate a strong ability to correct noisy labels and mitigate misleading supervision, further confirming its robustness.

\begin{table*}[htbp]
\centering
\scriptsize
\caption{Performance(ACLT, AILT, AILMT, AUCS, AUU, AUIS)  under various types of noise (rate 0.3) on NTU2012.}
\label{tab:mulperformance}
\begin{tabular}{lllccccccc}
\toprule
Dataset & Noise Type & Metrics & HGNN & S-model & APL & Backward & NRGNN & CGNN & OURS \\
\midrule
\multirow{18}{*}{NTU2012} & \multirow{6}{*}{Pair} 
& ACLT$\uparrow$ & 84.41 $\pm$ 3.99 & \textbf{85.19 $\pm$ 4.49} & 69.70 $\pm$ 4.27 & 15.21 $\pm$ 18.98 & 74.55 $\pm$ 2.40 & 74.36 $\pm$ 4.87 & 83.07 $\pm$ 2.34 \\
& & AILT$\uparrow$ & 58.82 $\pm$ 3.14 & 55.72 $\pm$ 5.38 & 45.40 $\pm$ 5.16 & 13.41 $\pm$ 16.63 & 58.73 $\pm$ 3.93 & 56.11 $\pm$ 5.86 & \textbf{67.16 $\pm$ 3.50} \\
& & AILMT$\downarrow$ & 30.37 $\pm$ 5.08 & 33.54 $\pm$ 10.05 & 38.51 $\pm$ 3.73 & \textbf{4.10 $\pm$ 2.68} & 19.89 $\pm$ 3.01 & 28.65 $\pm$ 5.06 & 18.10 $\pm$ 3.53 \\
& & AUCS$\uparrow$ & 74.72 $\pm$ 3.49 & 73.83 $\pm$ 3.99 & 62.13 $\pm$ 6.99 & 14.75 $\pm$ 17.47 & 73.93 $\pm$ 3.24 & 72.85 $\pm$ 5.41 & \textbf{78.84 $\pm$ 3.08} \\
& & AUU$\uparrow$ & 86.25 $\pm$ 5.93 & 86.25 $\pm$ 7.25 & 63.12 $\pm$ 23.93 & 18.12 $\pm$ 25.38 & 86.88 $\pm$ 6.22 & 86.25 $\pm$ 11.90 & \textbf{93.44 $\pm$ 5.20} \\
& & AUIS$\uparrow$ & 69.05 $\pm$ 4.58 & 66.99 $\pm$ 4.96 & 57.17 $\pm$ 6.93 & 14.41 $\pm$ 17.87 & 69.51 $\pm$ 3.28 & 67.00 $\pm$ 5.46 & \textbf{74.99 $\pm$ 3.47} \\
\cmidrule{2-10}
& \multirow{6}{*}{Uniform} 
& ACLT$\uparrow$ & 88.77 $\pm$ 2.25 & 89.47 $\pm$ 3.58 & 73.10 $\pm$ 5.15 & 50.61 $\pm$ 28.11 & 83.44 $\pm$ 1.98 & 79.58 $\pm$ 6.42 & \textbf{89.56 $\pm$ 1.37} \\
& & AILT$\uparrow$ & 79.85 $\pm$ 2.60 & 79.47 $\pm$ 2.52 & 56.49 $\pm$ 5.49 & 46.89 $\pm$ 26.18 & 74.60 $\pm$ 2.15 & 70.56 $\pm$ 5.39 & \textbf{82.23 $\pm$ 2.49} \\
& & AILMT$\downarrow$ & 1.64 $\pm$ 1.25 & 2.47 $\pm$ 2.88 & 6.91 $\pm$ 1.97 & 1.62 $\pm$ 0.83 & 1.27 $\pm$ 0.56 & 1.65 $\pm$ 0.82 & \textbf{0.64 $\pm$ 0.63} \\
& & AUCS$\uparrow$ & 85.06 $\pm$ 1.32 & 85.71 $\pm$ 1.48 & 69.94 $\pm$ 7.35 & 49.32 $\pm$ 27.63 & 83.72 $\pm$ 1.81 & 80.61 $\pm$ 5.58 & \textbf{86.89 $\pm$ 1.63} \\
& & AUU$\uparrow$ & 94.38 $\pm$ 4.84 & 96.25 $\pm$ 3.23 & 69.38 $\pm$ 21.54 & 55.31 $\pm$ 35.36 & 92.50 $\pm$ 2.64 & 94.69 $\pm$ 6.26 & \textbf{97.81 $\pm$ 3.31} \\
& & AUIS$\uparrow$ & 82.13 $\pm$ 1.55 & 82.60 $\pm$ 2.08 & 66.70 $\pm$ 8.47 & 48.36 $\pm$ 27.37 & 81.13 $\pm$ 2.42 & 76.66 $\pm$ 6.10 & \textbf{84.64 $\pm$ 1.71} \\
\cmidrule{2-10}
& \multirow{6}{*}{Random} 
& ACLT$\uparrow$ & 89.42 $\pm$ 2.13 & \textbf{90.60 $\pm$ 2.59} & 76.72 $\pm$ 3.37 & 48.75 $\pm$ 29.20 & 83.00 $\pm$ 1.30 & 77.91 $\pm$ 6.42 & 88.99 $\pm$ 2.15 \\
& & AILT$\uparrow$ & 78.94 $\pm$ 2.14 & 79.28 $\pm$ 3.46 & 57.99 $\pm$ 5.42 & 46.18 $\pm$ 26.68 & 71.62 $\pm$ 2.19 & 68.17 $\pm$ 9.35 & \textbf{80.56 $\pm$ 2.67} \\
& & AILMT$\downarrow$ & 1.87 $\pm$ 0.68 & 2.91 $\pm$ 3.04 & 8.12 $\pm$ 1.24 & 1.77 $\pm$ 1.24 & 1.37 $\pm$ 1.26 & 1.69 $\pm$ 0.79 & \textbf{0.68 $\pm$ 0.62} \\
& & AUCS$\uparrow$ & 84.56 $\pm$ 1.51 & 85.64 $\pm$ 0.75 & 72.25 $\pm$ 5.78 & 46.90 $\pm$ 28.73 & 82.62 $\pm$ 1.60 & 79.58 $\pm$ 3.95 & \textbf{86.21 $\pm$ 1.63} \\
& & AUU$\uparrow$ & 95.94 $\pm$ 5.11 & 95.94 $\pm$ 4.43 & 73.12 $\pm$ 21.86 & 55.31 $\pm$ 34.17 & 91.88 $\pm$ 3.95 & 95.00 $\pm$ 4.70 & \textbf{99.06 $\pm$ 2.11} \\
& & AUIS$\uparrow$ & 81.87 $\pm$ 2.66 & 82.78 $\pm$ 1.69 & 67.87 $\pm$ 7.62 & 45.63 $\pm$ 27.67 & 79.69 $\pm$ 2.38 & 76.24 $\pm$ 3.73 & \textbf{84.17 $\pm$ 2.66} \\
\bottomrule
\end{tabular}
\end{table*}

\subsubsection{Results of different Backbones}
To further evaluate the generality of HyperTrust across different architectures, we conduct experiments with different backbone architectures, where all methods are implemented using the same backbone in each setting. The results are reported in Tables \ref{tab:GAT} and \ref{tab:sage}: Table \ref{tab:GAT} presents the results with UniGNN~\cite{huang2021unignn} as the backbone, while Table \ref{tab:sage} reports the results with UniGAT~\cite{huang2021unignn} as the backbone. The best performance is highlighted in red. As shown in the tables, HyperTrust consistently achieves superior performance over competing methods in most cases across different backbones, further confirming its general effectiveness.
\begin{table*}[t]
\centering
\caption{Node classification performance (Accuracy (\%)$\pm$Std) on Cora, Cora-CA and NTU2012 under various types of noise (rate 0.3) using UniGNN as the backbone.}
\label{tab:GAT}
\begin{adjustbox}{max width=\textwidth}
\begin{tabular}{c|ccc|ccc|ccc}
  \toprule
  \multirow{2}{*}{\centering Dataset} 
  & \multicolumn{3}{c|}{CORA} 
  & \multicolumn{3}{c|}{Cora-CA}
  & \multicolumn{3}{c}{NTU2012}\\
   & pair & uniform & random 
   & pair & uniform & random
   & pair & uniform & random\\
  \hline
 UniGNN 
    & $66.95 \pm 3.96$ & $72.51 \pm 1.44$ & $72.17 \pm 1.31$
    & $70.80 \pm 2.66$ & $76.42 \pm 1.60$ & $76.61 \pm 1.69$
    & $74.86 \pm 2.97$ & $85.50 \pm 1.81$ & $85.83 \pm 1.15$ \\
  \hline
 S-model  
    & $67.79 \pm 3.84$ & $72.41 \pm 1.29$ & $72.81 \pm 1.31$
    & $71.10 \pm 1.57$ & $76.59 \pm 1.56$ & $76.84 \pm 1.35$
    & $75.34 \pm 4.14$ & $86.59 \pm 1.33$ & $86.54 \pm 1.37$ \\

 APL 
    & $67.46 \pm 5.59$ & $71.86 \pm 3.22$ & $72.50 \pm 2.38$
    & $70.54 \pm 3.80$ & $76.52 \pm 2.12$ & $76.39 \pm 2.88$
    & $60.07 \pm 6.50$ & $66.85 \pm 7.31$ & $70.09 \pm 7.74$ \\

 Backward 
    & $55.45 \pm 23.10$ & $72.83 \pm 2.36$ & $71.04 \pm 3.02$
    & $47.09 \pm 15.26$ & $68.12 \pm 7.90$ & $69.42 \pm 4.79$
    & $11.25 \pm 11.75$ & $49.20 \pm 25.61$ & $45.51 \pm 28.38$ \\

 NRGNN 
    & $63.85 \pm 2.45$ & $68.61 \pm 1.54$ & $69.70 \pm 1.11$
    & $64.86 \pm 2.14$ & $67.94 \pm 2.01$ & $68.76 \pm 1.20$
    & $74.17 \pm 3.21$ & $84.34 \pm 1.69$ & $83.46 \pm 1.55$ \\

 HyperTrust
    & \textcolor{red}{\textbf{$70.51 \pm 3.03$}} 
    & \textcolor{red}{\textbf{$74.13 \pm 1.14$}} 
    & \textcolor{red}{\textbf{$75.12 \pm 1.37$}}
    & \textcolor{red}{\textbf{$72.25 \pm 2.16$}} 
    & \textcolor{red}{\textbf{$78.04 \pm 1.50$}} 
    & \textcolor{red}{\textbf{$77.40 \pm 1.30$}}
    & \textcolor{red}{\textbf{$79.05 \pm 3.19$}} 
    & \textcolor{red}{\textbf{$88.36 \pm 1.46$}} 
    & \textcolor{red}{\textbf{$87.64 \pm 1.26$}} \\
  \bottomrule
\end{tabular}
\end{adjustbox}
\captionsetup{labelformat=empty}
\end{table*}

\begin{table*}[t]
\centering
\caption{Node classification performance (Accuracy (\%)$\pm$Std) on Cora, Cora-CA and NTU2012 under various types of noise (rate 0.3) using UniGAT as the backbone.}
\label{tab:sage}
\begin{adjustbox}{max width=\textwidth}
\begin{tabular}{c|ccc|ccc|ccc}
  \toprule
  \multirow{2}{*}{\centering Dataset} 
  & \multicolumn{3}{c|}{CORA} 
  & \multicolumn{3}{c|}{Cora-CA}
  & \multicolumn{3}{c}{NTU2012}\\
   & pair & uniform & random 
   & pair & uniform & random
   & pair & uniform & random\\
  \hline
 UniGAT    
    & $66.91 \pm 3.41$ & $72.35 \pm 2.74$ & $71.78 \pm 1.88$
    & $70.50 \pm 2.65$ & $76.66 \pm 1.26$ & $76.23 \pm 1.51$
    & $74.10 \pm 3.38$ & $85.72 \pm 1.62$ & $85.42 \pm 1.33$ \\
  \hline
 S-model  
    & $66.33 \pm 3.53$ & $72.62 \pm 2.10$ & $71.63 \pm 2.25$
    & $70.89 \pm 3.40$ & $76.10 \pm 1.40$ & $76.00 \pm 1.66$
    & $75.18 \pm 3.75$ & $85.70 \pm 1.76$ & $85.96 \pm 1.00$ \\

 APL   
    & $66.21 \pm 4.87$ & $73.03 \pm 2.24$ & $71.31 \pm 4.01$
    & $70.17 \pm 3.94$ & $76.24 \pm 1.66$ & $76.26 \pm 3.60$
    & $47.99 \pm 6.68$ & $59.46 \pm 8.21$ & $56.09 \pm 5.85$ \\

 Backward 
    & $48.96 \pm 20.19$ & $72.37 \pm 3.18$ & $66.28 \pm 10.70$
    & $51.92 \pm 10.62$ & $70.53 \pm 5.29$ & $69.13 \pm 4.71$
    & $5.88 \pm 5.73$ & $48.38 \pm 33.79$ & $47.36 \pm 20.24$ \\

 NRGNN  
    & $64.39 \pm 2.15$ & $68.48 \pm 1.29$ & $69.25 \pm 0.51$
    & $65.69 \pm 2.60$ & $68.12 \pm 1.64$ & $68.87 \pm 1.36$
    & $74.45 \pm 2.55$ & $84.53 \pm 1.49$ & $83.39 \pm 1.37$ \\

 HyperTrust 
    & \textcolor{red}{\textbf{$69.39 \pm 4.12$}} 
    & \textcolor{red}{\textbf{$74.39 \pm 0.92$}} 
    & \textcolor{red}{\textbf{$74.79 \pm 1.80$}}
    & \textcolor{red}{\textbf{$72.61 \pm 2.21$}} 
    & \textcolor{red}{\textbf{$78.72 \pm 1.47$}} 
    & \textcolor{red}{\textbf{$77.34 \pm 1.17$}}
    & \textcolor{red}{\textbf{$80.20 \pm 2.33$}} 
    & \textcolor{red}{\textbf{$87.77 \pm 1.28$}} 
    & \textcolor{red}{\textbf{$87.84 \pm 0.88$}} \\
  \bottomrule
\end{tabular}
\end{adjustbox}
\captionsetup{labelformat=empty}
\end{table*}

\subsection{ Impacts of Noisy Label Rate}
\begin{figure*}[h!]
\centering  
\subfigure[\textbf{Citeseer}]
{
\includegraphics[width=5.5cm,height =4.12cm]{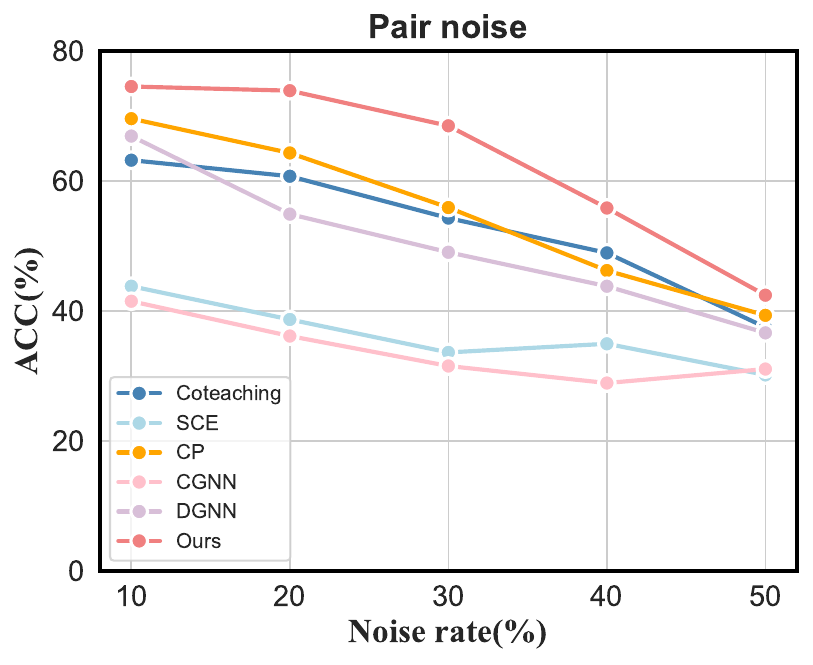}}
\subfigure[\textbf{Citeseer}]
{
\includegraphics[width=5.5cm,height =4.12cm]{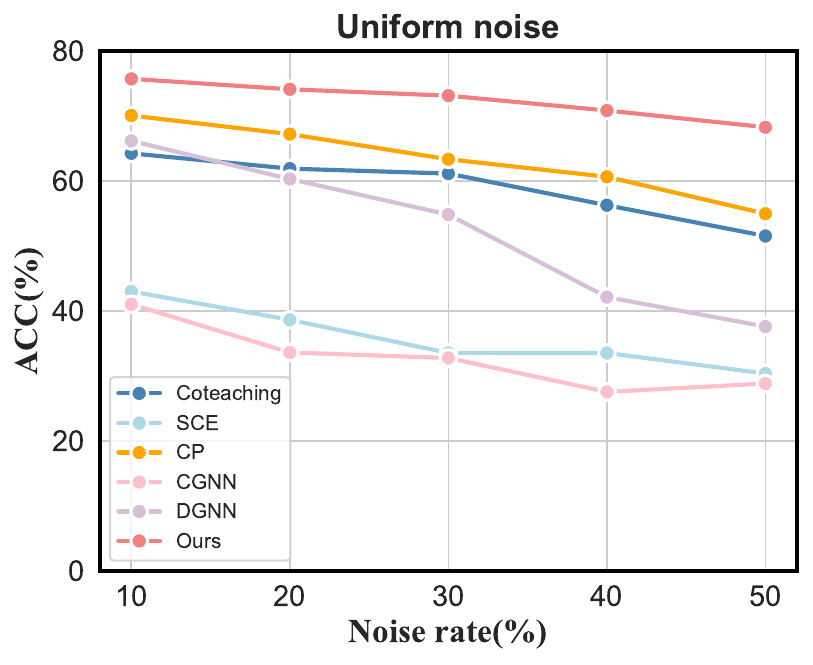}}
\subfigure[\textbf{Citeseer}]
{
\includegraphics[width=5.5cm,height =4.12cm]{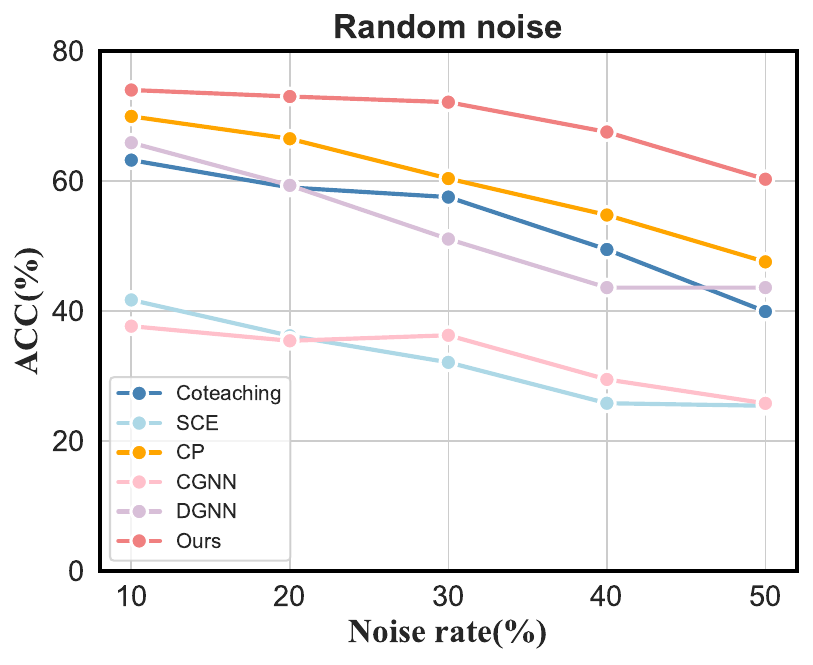}}
\subfigure[\textbf{Pubmed}]
{
\includegraphics[width=5.5cm,height =4.12cm]{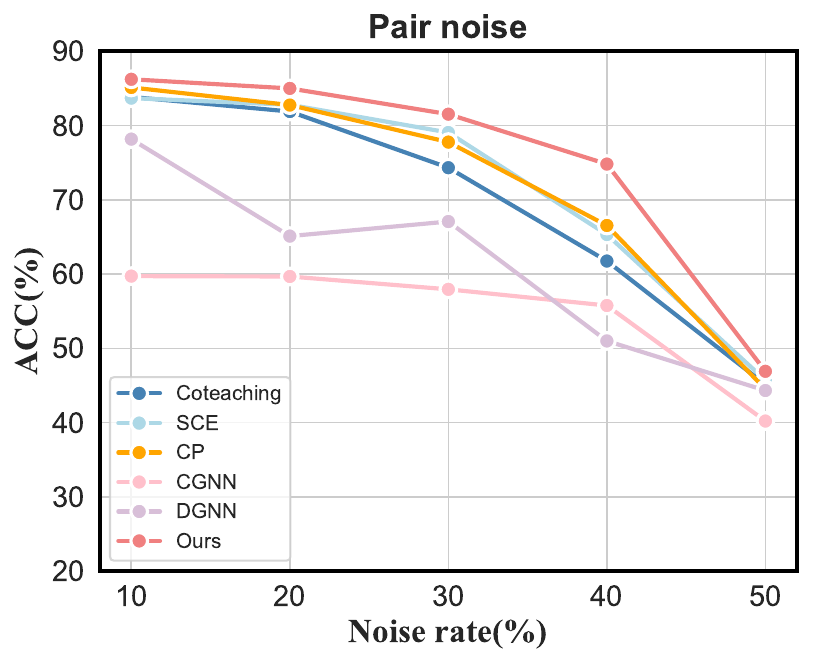}}
\subfigure[\textbf{Pubmed}]
{
\includegraphics[width=5.5cm,height =4.12cm]{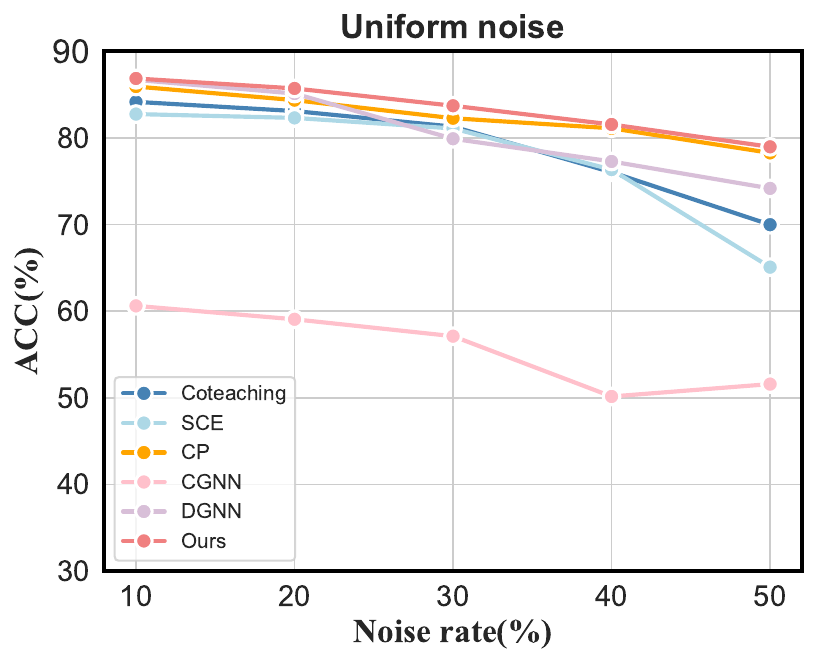}}
\subfigure[\textbf{Pubmed}]
{
\includegraphics[width=5.5cm,height =4.12cm]{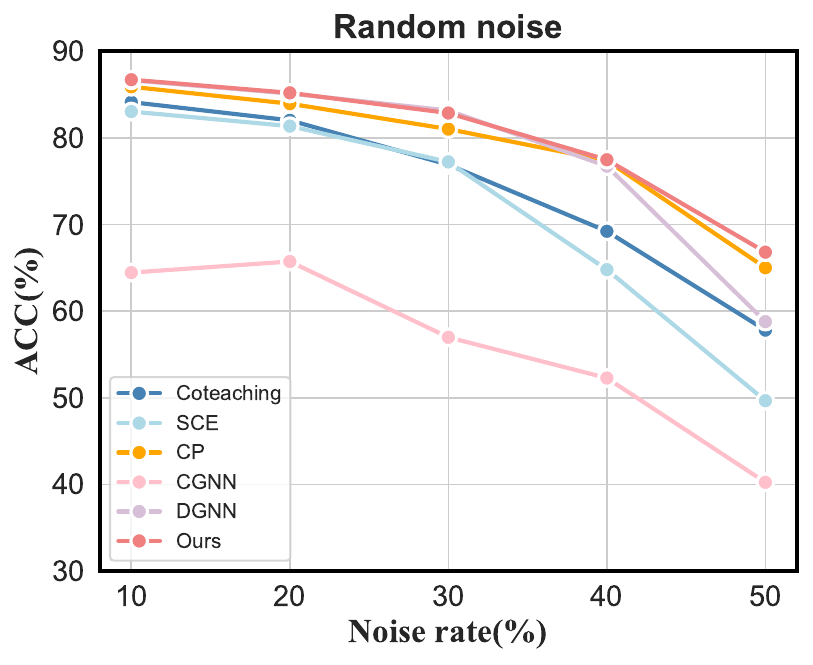}}
\caption{(a)(b)(c) show accuracy on Citeseer with various levels of label noise. (d)(e)(f) shows accuracy on Pubmed with various levels of label noise.}
\label{fig:noise rate}
\end{figure*}

To further evaluate the robustness of HyperTrust under different noise levels, we conducted experiments on the HyperTrust  with varying noise rates. As illustrated in Fig. \ref{fig:noise rate}, the accuracy trends of different methods on Citeseer and Pubmed are reported under three types of noise as the noise rate increases. From Fig. \ref{fig:noise rate}, we can draw the following observations:

\begin{itemize}
  \item As the noise rate increases, the accuracy of all methods generally decreases. Nevertheless, HyperTrust consistently achieves competitive  performance across datasets with different average degrees, indicating its strong robustness against noisy conditions.
  
  \item On the Citeseer dataset, under both uniform and random noise, the performance gap between HyperTrust and other methods becomes increasingly larger as the noise rate rises. This indicates that the robustness advantage of HyperTrust becomes increasingly pronounced as the noise rate increases.

\end{itemize}

\subsection{Ablation Study}
In this section, we perform ablation studies on multiple datasets to investigate the effectiveness of each  component in HyperTrust. The compared variants are defined as follows:

\begin{itemize}
\item[$\bullet$]
\textbf{w/o Aware:} This variant removes the Entropy-aware Hyperedge Trustworthiness Estimation module, meaning that no hyperedge estimation is performed before HyperedgeBoost and HyperedgePrune.

\item[$\bullet$]
\textbf{w/o Boost:} This variant removes the HyperedgeBoost module from HyperTrust and performs node classification only on the hypergraph pruned by HyperedgePrune.

\item[$\bullet$]
\textbf{w/o Prune:} This variant excludes the HyperedgePrune module and conducts node classification directly on $H^{Boost}$.
\end{itemize}

As reported in Table~\ref{tab:ab}, removing any of these modules leads to a noticeable drop in node classification performance, indicating that Hyperedge Trustworthiness Estimation , HyperedgeBoost, and HyperedgePrune all play important roles in capturing reliable interaction patterns. In particular,  Entropy-aware Hyperedge Trustworthiness Estimation helps identify informative hyperedges before subsequent edge enhancement and pruning operations, thereby reducing the influence of less relevant structures. Moreover, on dense datasets such as Pubmed, the variant using only HyperedgePrune  achieves better results than the variant using only HyperedgeBoost. This further demonstrates the importance of pruning noisy or redundant edges for improving model robustness.

\begin{table}[htp]
  \centering
  \caption{Ablation  Study on Cora,Citeseer and Pubmed.}
  \label{tab:ab}
  \begin{tabular}{c|c|c|c|c}
    \hline
    Dataset & Variants & Pair & Uniform & Random \\
    \hline

    \multirow{4}{*}{\centering Cora}
    & w/o Aware
    & $69.94 \pm 2.77$ 
    & $73.74 \pm 2.43$ 
    & $73.69 \pm 2.16$ \\
    
    & w/o Boost   
    & $64.33 \pm 4.09$ 
    & $70.01 \pm 2.50$ 
    & $69.38 \pm 3.25$ \\
    
    & w/o Prune  
    & $66.85 \pm 1.79$ 
    & $69.08 \pm 3.15$ 
    & $68.86 \pm 3.12$ \\
    
    & All
    & \textbf{70.43 $\pm$ 2.16} 
    & \textbf{74.75 $\pm$ 0.98} 
    & \textbf{75.06 $\pm$ 1.95} \\
    \hline

    \multirow{4}{*}{\centering Citeseer}
    & w/o Aware
    & $67.26 \pm 2.24$ 
    & $72.44 \pm 1.00$ 
    & $70.48 \pm 2.03$ \\
    
    & w/o Boost 
    & $64.92 \pm 2.25$ 
    & $69.02 \pm 1.44$ 
    & $68.75 \pm 2.02$ \\
    
    & w/o Prune    
    & $63.99 \pm 4.26$ 
    & $70.28 \pm 2.53$ 
    & $67.67 \pm 3.49$ \\
    
    & All 
    & \textbf{68.54 $\pm$ 2.67} 
    & \textbf{73.18 $\pm$ 0.90} 
    & \textbf{72.18 $\pm$ 2.03} \\
    \hline

    \multirow{4}{*}{\centering Pubmed}
    & w/o Aware 
    & $79.54 \pm 1.04$ 
    & $82.26 \pm 0.50$ 
    & $81.66 \pm 1.91$ \\
    
    & w/o Boost   
    & $78.87 \pm 0.95$ 
    & $81.12 \pm 0.94$ 
    & $80.67 \pm 1.20$ \\
    
    & w/o Prune  
    & $74.72 \pm 1.79$ 
    & $78.45 \pm 1.11$ 
    & $76.99 \pm 2.27$ \\
    
    & All  
    & \textbf{81.54 $\pm$ 1.20} 
    & \textbf{83.72 $\pm$ 0.75} 
    & \textbf{82.88 $\pm$ 1.72} \\
    \hline
  \end{tabular}
\end{table}

\subsection{Analysis of Computational Efficiency}
\begin{figure*}[htp]
\centering  
\subfigure[\textbf{Cora-CA(Pair)}]
{
\includegraphics[width=5.5cm,height =4.12
cm]{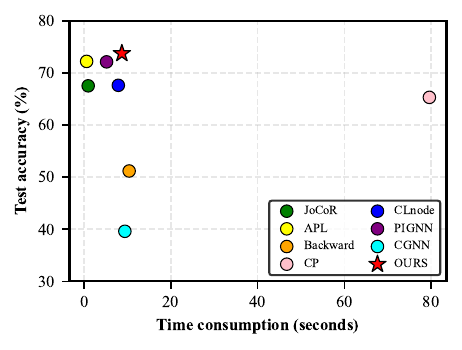}}
\subfigure[\textbf{Cora-CA(Uniform)}]
{
\includegraphics[width=5.5cm,height =4.12cm]{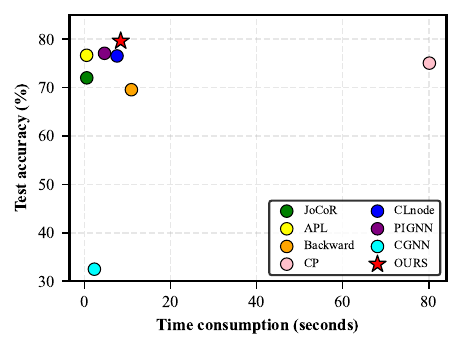}}
\subfigure[\textbf{Cora-CA(Random)}]
{
\includegraphics[width=5.5cm,height =4.12cm]{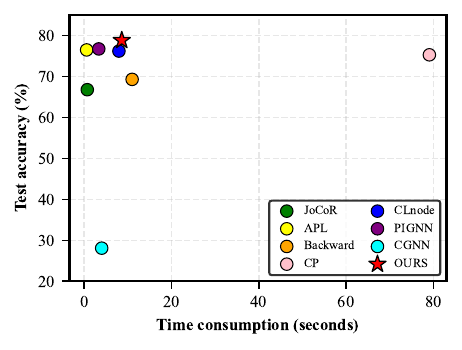}}
\subfigure[\textbf{NTU2012(Pair)}]
{
\includegraphics[width=5.5cm,height =4.12cm]{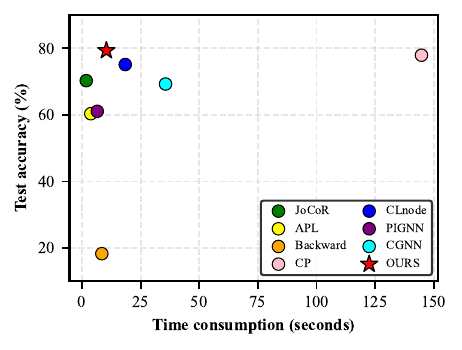}}
\subfigure[\textbf{NTU2012(Uniform)}]
{
\includegraphics[width=5.5cm,height =4.12cm]{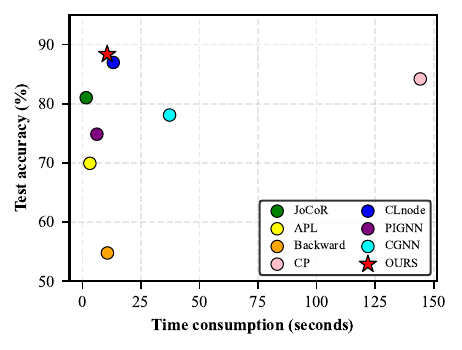}}
\subfigure[\textbf{NTU2012(Random)}]
{
\includegraphics[width=5.5cm,height =4.12cm]{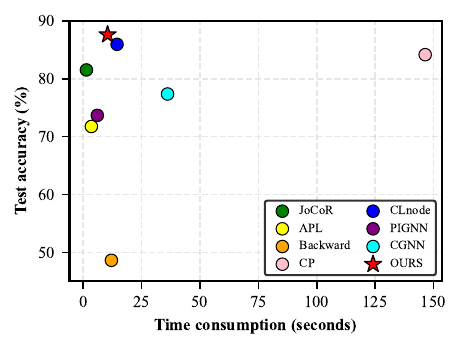}}
\caption{(a)–(c) show  accuracy and runtime on Cora-CA. (d)–(f)  show  accuracy and runtime on NTU2012.}
\label{runtime}
\end{figure*}

In this section, we record the runtime and test accuracy of different methods on Cora-CA and NTU2012 under 0.3 label noise. For each method, we performed 10 independent runs on each dataset. In each run, we recorded the average time required for the model to reach its best validation accuracy, and used this value as the total runtime of the corresponding method. The results are presented in Fig.~\ref{runtime}. As shown in the figure, the red nodes denote HyperTrust, which achieves a favorable trade-off between accuracy and computational efficiency on both datasets. By contrast, the competing methods either suffer from inferior accuracy or require substantially longer running time.

\subsection{Sensitivity Analysis of Hyper-parameters}
\begin{figure*}[htb]
\centering  
\subfigure[\textbf{Cora}]
{
\includegraphics[width=5.5cm,height =4.12cm]{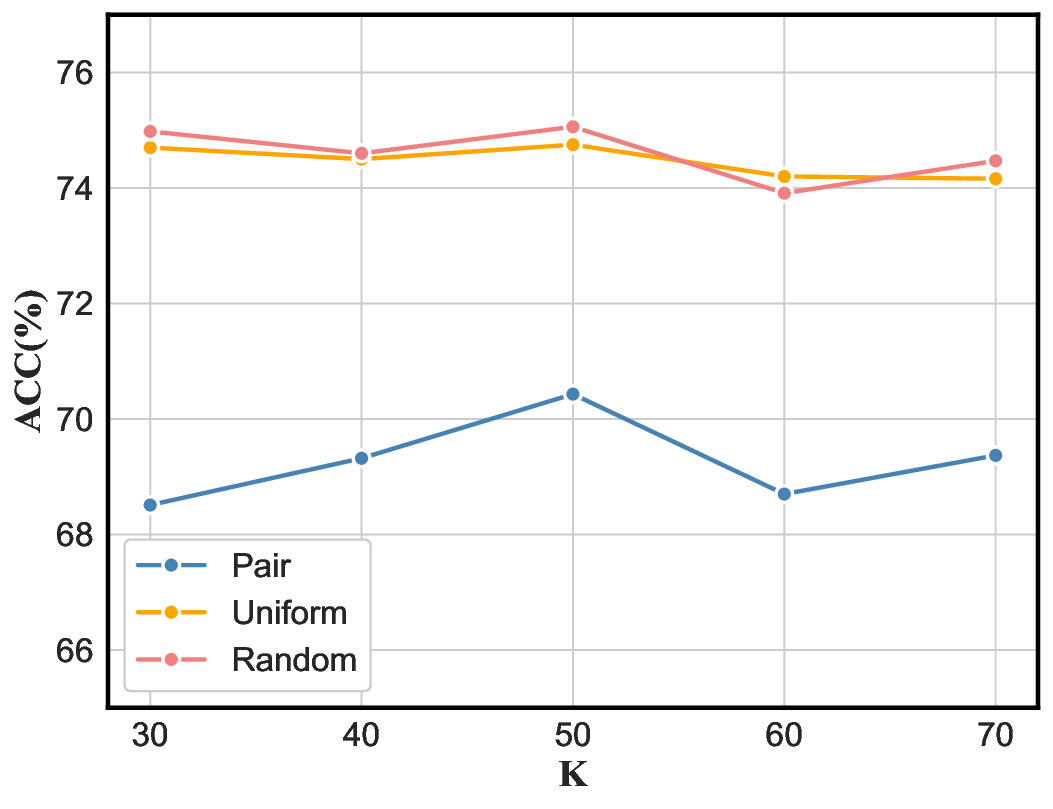}}
\subfigure[\textbf{ModelNet40}]
{
\includegraphics[width=5.5cm,height =4.12cm]{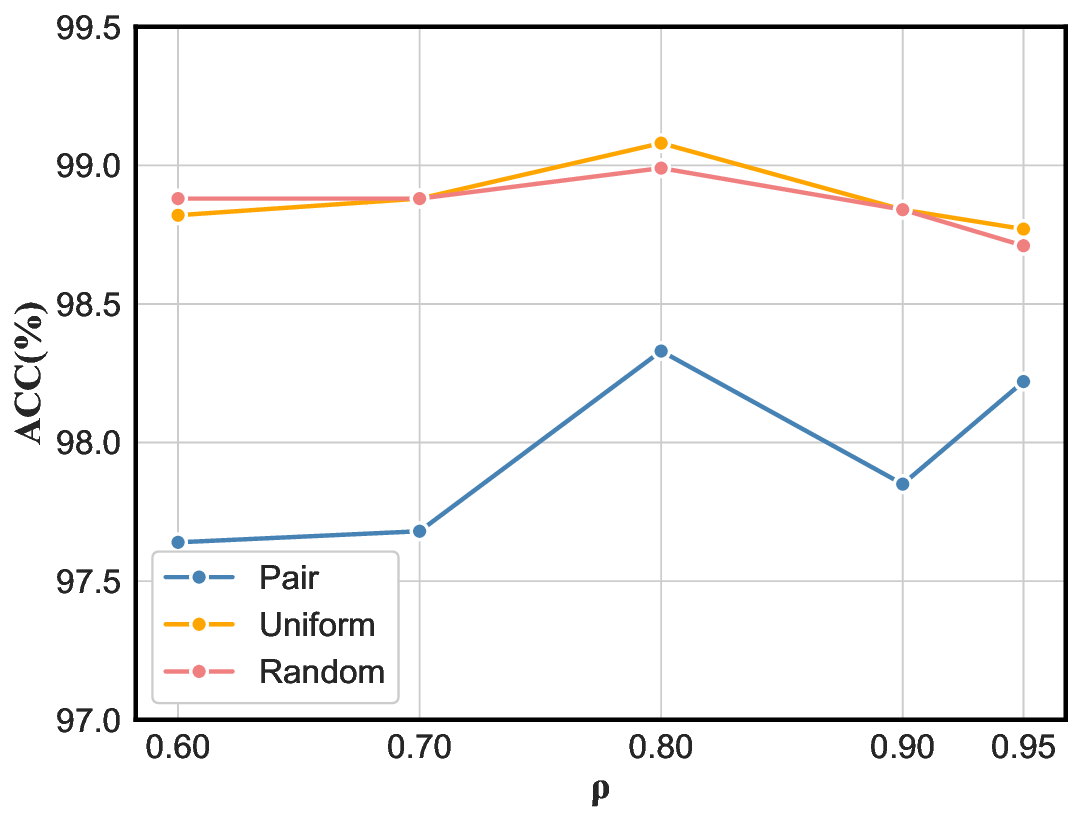}}
\subfigure[\textbf{DBLP}]
{
\includegraphics[width=5.5cm,height =4.12cm]{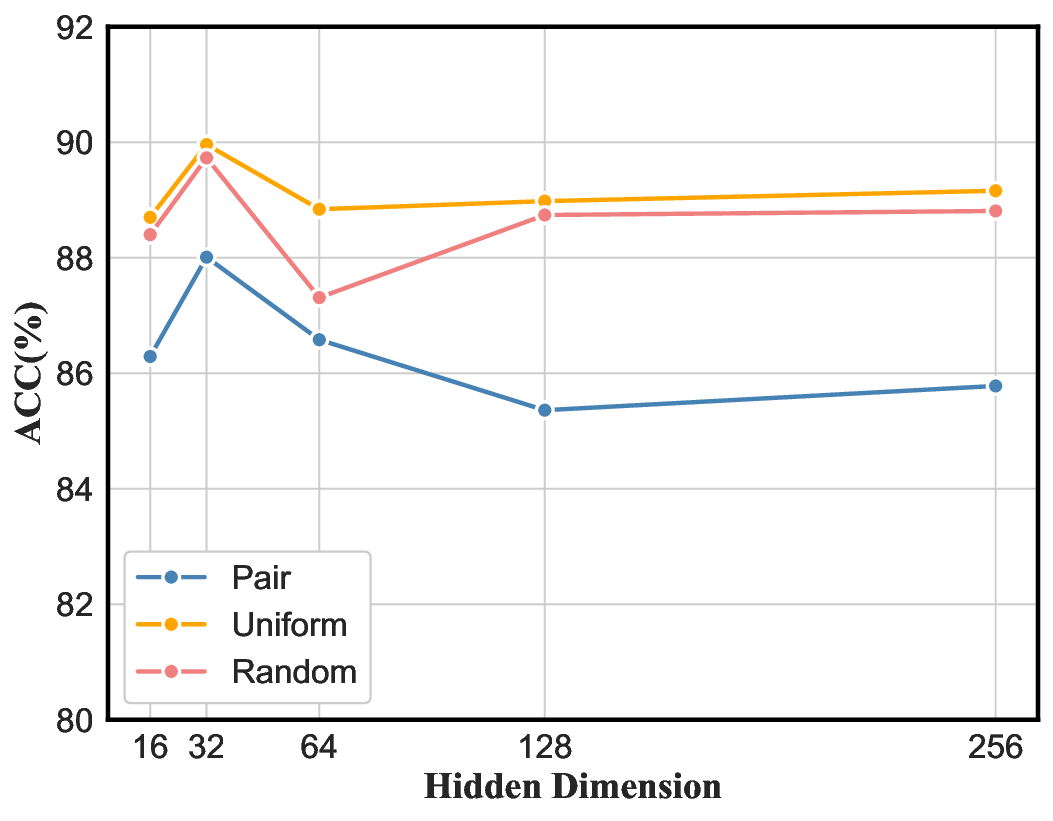}}
\caption{(a)–(c) show accuracy on Cora,ModelNet40, DBLP with various hyper-parameters  $K$, $\rho$, $Hidden Dimension$ and various noise types, respectively.}
\label{para}
\end{figure*}
In this section, we analyze the sensitivity of HyperTrust to key hyper-parameters in Fig.~\ref{para}, where $K$ denotes the number of selected trustworthy hyperedges for each unlabeled node, $\rho$ represents the pruning threshold, and the hidden dimension controls the size of node representations. \par

As shown in Figure~\ref{para}, HyperTrust is evaluated under pair, uniform, and random noise on Cora, ModelNet40, and DBLP, respectively. In Fig.~\ref{para}(a), the performance on Cora remains relatively stable as $K$ varies, and the best results are generally achieved around $K=50$. This indicates that selecting a moderate number of trustworthy hyperedges is beneficial for enhancing reliable high-order propagation. \par

Fig.~\ref{para}(b) shows the effect of $\rho$ on ModelNet40. The performance first improves and then slightly decreases as $\rho$ increases, with the best performance obtained around $\rho=0.8$. This suggests that an appropriate pruning threshold can effectively suppress noisy incidence relations, while an overly large threshold may remove useful structural information. \par

Fig.~\ref{para}(c) reports the influence of the hidden dimension on DBLP. The model achieves competitive performance across different hidden dimensions, with the best results generally appearing at 32. Overall, HyperTrust shows stable performance under different hyper-parameter settings, demonstrating its robustness to hyper-parameter variations.
\subsection{Case Study}
To further validate the effectiveness of HyperTrust, we conduct a statistical analysis of the results produced by its three key procedures after training. Table~\ref{tab:strategy_results}  reports the statistical results of HyperTrust under different noise types. Specifically, TN denotes the proportion of noisy hyperedges among the trustworthy hyperedges selected by HyperTrust. Here, a hyperedge is regarded as noisy if it contains at least one noisy node. TN2 further measures the proportion of selected trustworthy hyperedges that contain two or more noisy nodes. UTN denotes the proportion of noisy hyperedges among the hyperedges identified as untrustworthy. BT measures the proportion of boosted hyperedge-node associations in HyperedgeBoost that can provide correct supervision signals for unlabeled nodes, while PT denotes the proportion of pruned nodes in HyperedgePrune that are indeed noisy nodes.

As shown in Table~\ref{tab:strategy_results} , TN is consistently much lower than UTN across all datasets and noise types. This indicates that the Hyperedge Trustworthiness Estimation module in HyperTrust can effectively distinguish relatively clean hyperedges from noisy ones. In addition, TN2 is also substantially lower than TN in all cases, suggesting that although some hyperedges selected as trustworthy may still contain noisy nodes, severe contamination involving multiple noisy nodes is relatively rare. Notably, TN and TN2 are relatively high on ModelNet40, mainly because the number of truly clean hyperedges becomes limited after label noise is injected, making trustworthy hyperedge selection particularly challenging.

The BT and PT results further validate the effectiveness of HyperedgeBoost and HyperedgePrune. Overall, the BT values remain high, indicating that the enhanced hyperedge-node associations can usually provide reliable supervision signals for unlabeled nodes. Meanwhile, the PT results show that the pruning operation can accurately remove truly noisy nodes, especially under the random noise setting. These results demonstrate that HyperTrust can not only effectively select high-quality clean hyperedges, but also optimize the hypergraph structure by adding reliable supervisory associations and pruning noisy structures.

\begin{table}[t] 
\centering 
\caption{Case study results (\%)  on Pubmed, DBLP-CA, and ModelNet40 under different noise types.}
\label{tab:strategy_results} 
\resizebox{\linewidth}{!}{ \begin{tabular}{llccccc} 
\toprule Dataset & Noise Type & TN$\downarrow$ & TN2$\downarrow$ & UTN$\uparrow$ & BT$\uparrow$ & PT$\uparrow$ \\ \midrule \multirow{3}{*}{Pubmed} & Pair & 53.52 & 16.75 & 76.78 & 79.72 & 73.33 \\ & Uniform & 49.38 & 12.68 & 77.73 & 78.24 & 83.33 \\ & Random & 50.52 & 15.55 & 78.41 & 78.02 & 89.09 \\ 
\midrule 
\multirow{3}{*}{DBLP-CA} & Pair & 45.25 & 14.10 & 81.11 & 80.65 & 88.89 \\ & Uniform & 43.78 & 11.96 & 78.54 & 80.65 & 85.29 \\ & Random & 44.63 & 12.32 & 79.43 & 77.76 & 93.55 \\ 
\midrule
\multirow{3}{*}{ModelNet40} & Pair & 69.35 & 35.33 & 85.76 & 85.40 & 86.09 \\ & Uniform & 72.45 & 41.40 & 86.01 & 88.27 & 89.56 \\ & Random & 68.98 & 36.23 & 85.87 & 83.70 & 97.06 \\ \bottomrule \end{tabular} } \end{table}

\section{Conclusion}
In this paper, we presented a systematic study of hypergraph node classification under label noise, a practical yet underexplored problem in hypergraph learning. We first constructed a unified benchmark by adapting representative LLN and GLN methods to hypergraphs, and empirically revealed that existing robust learning strategies are insufficient for handling noisy supervision in higher-order relational structures. To address this challenge, we proposed HyperTrust, a robust hypergraph learning framework. Specifically, HyperTrust estimates  hyperedge trustworthiness through a pretraining-based entropy-aware strategy, enhances trustworthy supervision via the HyperedgeBoost module, and suppresses noisy information propagation through the HyperedgePrune module. Extensive experiments under diverse noisy-label settings demonstrate that HyperTrust  outperforms adapted robust-learning baselines and strong hypergraph neural network backbones. Theoretical analysis further supports the effectiveness of our design in reducing the adverse impact of label noise on hypergraph message passing. Overall, our work establishes a new benchmark and provides an effective framework for robust hypergraph learning, offering a foundation for future research on hypergraph learning with label noise.
\bibliographystyle{IEEEtran}
\bibliography{IEEEabrv,refer}

\newpage

 




\clearpage
\setcounter{page}{1}
\appendices
\section{The pseudo-code of the HyperTrust}
\label{appa}
In this section, we present the pseudo-code of HyperTrust as follows:

\begin{algorithm}[htp] 
\caption{The algorithm of HyperTrust} 
\label{alg:hypertrust} 
\renewcommand{\algorithmicrequire}{\textbf{Input:}}
\renewcommand{\algorithmicensure}{\textbf{Output:}}
\begin{algorithmic}[1]
 \REQUIRE Hypergraph $\mathcal{H}=(\mathcal{V},\mathcal{E})$ with incidence matrix $H$, node feature matrix $X$,  noisy labels $Y_L^N$.
 \ENSURE Final prediction $Y^{Final}$.
\STATE Pretrain a HGNN encoder according to Eqs.~\ref{2}--~\ref{5}. \STATE Obtain node embeddings $Z$ and pseudo-labels $\hat{Y}$. 
\STATE Construct mixed labels $\tilde{Y}$ according to Eq.~\ref{6}. 
\FOR{each hyperedge $e \in \mathcal{E}$} 
\STATE Compute its label distribution entropy by Eqs.~\ref{7}--~\ref{8}. \ENDFOR
\STATE Divide hyperedges into $\mathcal{E}_{trust}$ and $\mathcal{E}_{untrust}$ according to Eq.~\ref{9} and Eq.~\ref{10}.
\STATE Calculate $H^{Boost}$ according to Eq.~\ref{eq:hgg_hyperedge_embedding}--Eq.~\ref{eq:hgg_boost_incidence}.
\STATE Calculate $H^{Prune}$ according to Eq.~\ref{eq:hgg_class_proto}--Eq.~\ref{eq:hgg_prune_incidence}.
\FOR{$t=0; t<epochs; t=t+1$}
\STATE Obtain $Y^{Boost}$ with $f_{\theta}$ according to Eq.~\ref{eq:hgg_boost_prediction}.
\STATE Obtain $Y^{Prune}$ with $f_{\theta}$ according to Eq.~\ref{eq:hgg_prune_prediction}.
\STATE Generate final prediction $Y^{Final}$ according to Eq.~\ref{eq:hgg_fusion}.\STATE Update parameters by minimizing Eq.~\ref{21}.
\ENDFOR
\RETURN $Y^{Final}$.
\end{algorithmic}
\end{algorithm}
\section{Proof of Theorem~\ref{thm:noise-margin}}
\label{appb}

\begin{Proof}
For any incorrect class $r\neq c$, by the definition of $m_i(P)$, we have
\begin{equation}
m_i(P)_c-m_i(P)_r
=
\sum_{j=1}^n P_{ij}
\left(
\mathbf{1}(y_j^N=c)-\mathbf{1}(y_j^N=r)
\right).
\end{equation}
Under the uniform label noise model, for $ s\neq y_j$ 
\begin{equation}
\mathbb{P}(y_j^N=y_j)=1-\epsilon,
\qquad
\mathbb{P}(y_j^N=s)=\frac{\epsilon}{C-1}.
\end{equation}
Therefore, for the term inside the summation, we have
\begin{align}
&\mathbb{E}\left[
\mathbf{1}(y_j^N=c)-\mathbf{1}(y_j^N=r)
\mid Y
\right] \nonumber\\
&\quad =
\begin{cases}
1-\epsilon-\dfrac{\epsilon}{C-1}, & y_j=c,\\[2mm]
-\left(1-\epsilon-\dfrac{\epsilon}{C-1}\right), & y_j=r,\\[2mm]
0, & y_j\notin\{c,r\}.
\end{cases}
\end{align}
Since $\lambda=1-\epsilon-\frac{\epsilon}{C-1}$, taking expectation gives
\begin{align}
&\mathbb{E}\left[m_i(P)_c-m_i(P)_r\mid Y\right] \nonumber\\
&=
\sum_{j=1}^n P_{ij}
\mathbb{E}\left[
\mathbf{1}(y_j^N=c)-\mathbf{1}(y_j^N=r)
\mid Y
\right] \nonumber\\
&=
\lambda\sum_{j:y_j=c}P_{ij}
-
\lambda\sum_{j:y_j=r}P_{ij} \nonumber\\
&=
\lambda\left(S_i^+(P)-S_{i,r}^-(P)\right).
\end{align}
This completes the proof.
\end{Proof}
More generally, let
$\mathbf{T}\in[0,1]^{C\times C}$ be a class-conditional label
transition matrix, where
\begin{equation}
T_{s,t}
=
\Pr(y_j^N=t\mid y_j=s).
\label{eq:transition-matrix}
\end{equation}
For any fixed propagation matrix $P$, the expected pairwise margin
under a general class-conditional noise model satisfies
\begin{equation}
\mathbb{E}
\left[
m_i(P)_c-m_i(P)_r
\mid Y
\right]
=
\sum_{s=1}^{C}
S_{i,s}(P)
\left(
T_{s,c}-T_{s,r}
\right),
\label{eq:general-transition-margin}
\end{equation}
where
\begin{equation}
S_{i,s}(P)
=
\sum_{j:y_j=s}P_{ij}.
\end{equation}
Therefore, pair noise and random noise can be analyzed by
substituting their corresponding transition matrices into
Eq.~\eqref{eq:general-transition-margin}.

\section{Proof of Theorem~\ref{prop:fusion}}
\label{appc}

\subsection{Proof of Lemma~\ref{lem:add}}
\begin{Proof}
The original propagation weights are scaled by $1-\eta_i$. Since the added hyperedges have purity at least $q_\star$ and are aligned with class $c$, at least $\eta_i q_\star$ weights contribute to the true class, while at most $\eta_i(1-q_\star)$ weights contribute to any wrong class. Hence,
\begin{equation}
S_i^+(P^{\mathrm{add}})
\ge
(1-\eta_i)S_i^+(P)+\eta_i q_\star,
\end{equation}
and
\begin{equation}
S_{i,r}^-(P^{\mathrm{add}})
\le
(1-\eta_i)S_{i,r}^-(P)+\eta_i(1-q_\star).
\end{equation}
Subtracting the two inequalities and taking the minimum over $r\neq c$ gives the result.
\end{Proof}
\subsection{Proof of Lemma~\ref{thm:drop}}
\begin{Proof}
After pruning and renormalization, we have
\begin{equation}
S_i^+(P^{\mathrm{Prune}})
=
\frac{S_i^+(P)-a_i}{1-\omega_i},
\end{equation}
\begin{equation}
    S_{i,r}^-(P^{\mathrm{Prune}})
=
\frac{S_{i,r}^-(P)-\beta_{i,r}}{1-\omega_i}.
\end{equation}
Since the strongest competing class remains $r^\star$ after pruning, the margin after pruning can be written as
\begin{align}
\Gamma_i(P^{\mathrm{Prune}})
&=
S_i^+(P^{\mathrm{Prune}})
-
S_{i,r^\star}^-(P^{\mathrm{Prune}}) \\
&=
\frac{
S_i^+(P)-S_{i,r^\star}^-(P)
+\beta_{i,r^\star}-a_i
}{1-\omega_i} \\
&=
\frac{
\Gamma_i(P)+\beta_{i,r^\star}-a_i
}{1-\omega_i}.
\end{align}
Because $\omega_i\in[0,1)$, we have $1-\omega_i>0$. Moreover, by assumption,
\begin{equation}
\beta_{i,r^\star}-a_i>0,
\end{equation}
and
\begin{equation}
\Gamma_i(P)\geq 0.
\end{equation}
Therefore,
\begin{equation}
\Gamma_i(P^{\mathrm{Prune}})
=
\frac{
\Gamma_i(P)+\beta_{i,r^\star}-a_i
}{1-\omega_i}
>
\Gamma_i(P),
\end{equation}
which completes the proof.
\end{Proof}
\subsection{Proof of Theorem~\ref{prop:fusion}}

\begin{Proof}
Fix a node $v_i$ and let its true label be $y_i$. For any wrong class
$r\neq y_i$, since the fused surrogate propagation matrix is defined as
\[
P^{\mathrm{Fuse}}
=
\frac{1}{2}( P^{\mathrm{Boost}}+P^{\mathrm{Prune}}),
\]
the corresponding true-class and wrong-class scores are also linear
combinations of the two scores. Hence,
\[
S_i^+(P^{\mathrm{Fuse}})
=
\frac{1}{2}(S_i^+(P^{\mathrm{Boost}})
+
S_i^+(P^{\mathrm{Prune}})),
\]
and
\[
S_{i,r}^-(P^{\mathrm{Fuse}})
=
\frac{1}{2} (S_{i,r}^-(P^{\mathrm{Boost}})
+
S_{i,r}^-(P^{\mathrm{Prune}})).
\]
Therefore,
\[
\begin{aligned}
&S_i^+(P^{\mathrm{Fuse}})-S_{i,r}^-(P^{\mathrm{Fuse}})\\
&=
\frac{1}{2}\big(S_i^+(P^{\mathrm{Boost}})
-S_{i,r}^-(P^{\mathrm{Boost}})\big) \\
&\quad+
\frac{1}{2}\big(S_i^+(P^{\mathrm{Prune}})
-S_{i,r}^-(P^{\mathrm{Prune}})\big).\\
\end{aligned}
\]
Taking the minimum over all $r\neq y_i$, we obtain
\[
\begin{aligned}
\Gamma_i(P^{\mathrm{Fuse}}) &=\min_{r\neq y_i}
\big\{
S_i^+(P^{\mathrm{Fuse}})-S_{i,r}^-(P^{\mathrm{Fuse}})
\big\}\\
&=
\min_{r\neq y_i}
\big\{
\frac{1}{2}\big(S_i^+(P^{\mathrm{Boost}})
-S_{i,r}^-(P^{\mathrm{Boost}})\big) \\
&\quad+
\frac{1}{2}\big(S_i^+(P^{\mathrm{Prune}})
-S_{i,r}^-(P^{\mathrm{Prune}})\big)
\big\} \\
&\ge
\frac{1}{2}\min_{r\neq y_i}(S_i^+(P^{\mathrm{Boost}})
-S_{i,r}^-(P^{\mathrm{Boost}})\big)\\
&\quad+
\frac{1}{2}\min_{r\neq y_i}\big(S_i^+(P^{\mathrm{Prune}})
-S_{i,r}^-(P^{\mathrm{Prune}})\big)\\
&=
\frac{1}{2}\Gamma_i(P^{\mathrm{Boost}})
+
\frac{1}{2}\Gamma_i(P^{\mathrm{Prune}})
\\&> \Gamma_i(P).
\end{aligned}
\]
This proves the desired result.
\end{Proof}

\section{Adaptation of Baselines to Hypergraph Learning}
\label{C}
  For fair comparison, we adapt all baselines originally designed for ordinary graphs or non-graph data to hypergraph
  node classification. Given a hypergraph
  $\mathcal{H}=(\mathcal{V},\mathcal{E})$, we represent its structure by the incidence matrix
  $H\in\{0,1\}^{|\mathcal{V}|\times |\mathcal{E}|}$, which is stored in the code as a COO-style
  $\texttt{edge\_index}=[v,e]$ containing node--hyperedge incidences. Unless otherwise specified, the original MLP/GCN/
  GNN backbone is replaced with an HGNN encoder:
  \[
      X' = D_v^{-\frac{1}{2}} H W_e D_e^{-1} H^\top D_v^{-\frac{1}{2}} X,
  \]
  where $D_v$ and $D_e$ denote the node-degree and hyperedge-degree matrices, respectively. This converts ordinary
  pairwise message passing into node--hyperedge--node high-order aggregation.

  \begin{itemize}
      \item \textbf{S-model.}
      We keep the learnable label-transition mechanism of S-model and replace its classifier with HGNN. The HGNN first
  predicts $p_\theta(y\mid x,H)$, which is then transformed by a learnable noise transition matrix $T$:
      \[
          \tilde{p}(y\mid x,H)=p_\theta(y\mid x,H)T .
      \]
      The adapted distribution is supervised by noisy labels.

      \item \textbf{Co-teaching.}
      We instantiate the two peer networks in Co-teaching as two HGNNs. Each HGNN selects small-loss samples according
  to its own per-node cross-entropy loss, and the selected samples are exchanged to update the other HGNN.

      \item \textbf{JoCoR.}
      JoCoR is adapted as two HGNN classifiers trained with a joint loss:
     
      \[
       \begin{aligned}
          \mathcal{L}_{\mathrm{JoCoR}}
          &=(1-\lambda)\left(\ell_{\mathrm{CE}}^{(1)}+\ell_{\mathrm{CE}}^{(2)}\right)\\
         & +\lambda\left(\mathrm{KL}(p_1\|p_2)+\mathrm{KL}(p_2\|p_1)\right).
          \end{aligned}
      \]
      Here, $p_1$ and $p_2$ denote the predictive class distributions produced by the two HGNN classifiers for the
  same node, and
        $\ell_{\mathrm{CE}}^{(1)}$ and $\ell_{\mathrm{CE}}^{(2)}$ are the cross-entropy losses between their predictions
  and the observed noisy label. During training,
  JoCoR computes this joint loss for each training node and keeps the small-loss nodes for parameter updates, since
  these nodes are more likely to be correctly labeled.
      \item \textbf{APL and SCE.}
        For APL and SCE, we retain their noise-robust loss functions and replace the original feature encoder with HGNN,
  so that node representations are learned from the hypergraph incidence structure. In APL, the training objective
  combines normalized cross entropy (NCE) and reverse cross entropy (RCE):
        \[
            \mathcal{L}_{\mathrm{APL}}
            = \alpha \mathcal{L}_{\mathrm{NCE}}
            + \beta \mathcal{L}_{\mathrm{RCE}} .
        \]
        Here, $\mathcal{L}_{\mathrm{NCE}}$ is a normalized version of the standard cross-entropy loss, which reduces the
  dominance of overly confident noisy samples, while $\mathcal{L}_{\mathrm{RCE}}$ reverses the roles of the predicted
  distribution and the label distribution to improve robustness against corrupted labels. The coefficients $\alpha$ and
  $\beta$ balance the two terms.

        Similarly, SCE is adapted by applying its symmetric loss to the HGNN outputs:
        \[
            \mathcal{L}_{\mathrm{SCE}}
            = \alpha \mathcal{L}_{\mathrm{CE}}
            + \beta \mathcal{L}_{\mathrm{RCE}} .
        \]
        In this objective, $\mathcal{L}_{\mathrm{CE}}$ denotes the standard cross-entropy loss between the HGNN
  prediction and the observed noisy label, which preserves the fitting ability of supervised learning. The reverse
  cross-entropy term $\mathcal{L}_{\mathrm{RCE}}$ acts as a noise-robust regularizer by penalizing inconsistent
  predictions in a symmetric manner. Thus, APL and SCE are transferred to hypergraph learning mainly by replacing the
  classifier backbone with HGNN while keeping their original robust-loss principles.

     \item \textbf{Forward and Backward correction.}
        For Forward and Backward correction, we first pretrain a hypergraph classifier on the noisy training labels and
  use its high-confidence predictions to estimate the label-transition matrix $C$. Each entry $C_{ij}$ represents the
  probability that a clean label $i$ is observed as a noisy label $j$. Let $\tilde{y}$ denote the one-hot vector of the
  observed noisy label, and let $p_\theta(y\mid x,H)$ denote the predicted clean-label distribution produced by the
  hypergraph classifier for node $x$ under the hypergraph incidence structure $H$.

        Forward correction modifies the model prediction before computing the loss:
        \[
            \mathcal{L}_{\mathrm{forward}}
            =-\tilde{y}^{\top}\log\left(p_\theta(y\mid x,H)C\right).
        \]
        In this formulation, $p_\theta(y\mid x,H)C$ maps the predicted clean-label distribution to the noisy-label space
  through the estimated transition matrix $C$. Therefore, the model is trained by matching the transformed prediction
  with the observed noisy label.

        Backward correction instead modifies the loss using the inverse transition process:
        \[
            \mathcal{L}_{\mathrm{backward}}
            =-\left(\tilde{y}^{\top}C^{\dagger}\right)\log p_\theta(y\mid x,H),
        \]
        where $C^{\dagger}$ is the pseudo-inverse of $C$. The term $\tilde{y}^{\top}C^{\dagger}$ can be regarded as a
  corrected supervision signal that compensates for label corruption before applying the cross-entropy loss. In our
  hypergraph adaptation, the transition matrix is estimated from a pretrained hypergraph classifier, and the final model
  is trained on the hypergraph structure with the corresponding forward or backward corrected objective.

      \item \textbf{NRGNN.}
      We replace edge estimation in NRGNN with node--hyperedge incidence estimation. An HGNN estimates node
  representations, hyperedge representations are computed by averaging incident node representations, and the score of a
  candidate incidence $(v,e)$ is given by their inner product. Candidate incidences are generated from training nodes
  and high-confidence unlabeled nodes, and the model jointly optimizes classification, incidence reconstruction, and
  pseudo-label consistency losses.

    \item \textbf{CP.}
        CP is adapted by introducing community-aware supervision into hypergraph node classification. Specifically, we
  first run HyperNode2Vec on the node--hyperedge bipartite structure induced by the hypergraph incidence matrix $H$. The
  random walks alternate between nodes and hyperedges, so the learned node embeddings encode high-order hypergraph
  connectivity. Then, KMeans is applied to these embeddings to obtain community pseudo-labels for all nodes.

        Based on these community pseudo-labels, the HGNN is trained with two prediction heads: one head predicts the
  original class label, and the other predicts the community assignment. The overall objective is
        \[
            \mathcal{L}_{\mathrm{CP}}
            =
            \mathcal{L}_{\mathrm{cls}}
            +
            \lambda \mathcal{L}_{\mathrm{cluster}} .
        \]
        Here, $\mathcal{L}_{\mathrm{cls}}$ is the standard classification loss between the HGNN class prediction and the
  observed noisy label, while $\mathcal{L}_{\mathrm{cluster}}$ is the cross-entropy loss between the community
  prediction and the KMeans-generated community pseudo-label. The coefficient $\lambda$ controls the strength of the
  community regularization. In this way, CP is transferred to hypergraphs by replacing ordinary graph-based community
  information with communities discovered from the node--hyperedge bipartite structure.

      \item \textbf{CLNode.}
      CLNode is adapted as a hypergraph curriculum-learning method. A pretrained HGNN provides node embeddings. Local
  difficulty is measured by the entropy of labels in the shared-hyperedge neighborhood, while global difficulty is
  measured by similarity between node embeddings and class prototypes. Training nodes are then introduced from easy to
  hard.

      \item \textbf{PIGNN.}
      PIGNN is adapted by constructing a hypergraph-induced node context matrix
      \[
          A_H=\mathbb{I}(HH^\top>0),
      \]
      where two nodes are context-related if they share at least one hyperedge. HGNN embeddings are decoded by an inner-
  product decoder to reconstruct this context matrix, and the reconstruction loss is combined with the node
  classification loss.

      \item \textbf{CGNN.}
      CGNN is implemented with HGNN and hypergraph contrastive augmentation. We construct two views by randomly
  dropping node--hyperedge incidences and node features, respectively, and enforce representation consistency with a
  contrastive loss. The model also uses supervised classification loss and, after warm-up, corrects noisy labels using
  majority labels in shared-hyperedge neighborhoods and representation similarity.

      \item \textbf{DGNN.}
      DGNN is adapted by replacing the graph encoder with HGNN. After HGNN pretraining, the transition matrix $C$ is
  estimated from high-confidence class representatives, and the main HGNN is trained with the backward-corrected loss.
  Thus, the noise-transition and loss-correction mechanisms are preserved, while structural learning is performed on the
  hypergraph incidence structure.
  \end{itemize}

\section{Limitations}
Although HyperTrust shows promising performance in hypergraph learning under label noise, it still has several limitations. First, our method is mainly based on the homophily assumption that nodes connected by the same hyperedge are likely to share similar semantic labels. Under this assumption, low-entropy hyperedges can be regarded as more trustworthy structures for message propagation. However, in heterophilic hypergraphs, nodes within the same hyperedge may naturally belong to different classes. In such cases, the entropy-based trustworthiness estimation may become less reliable, and some informative high-order relations may be mistakenly regarded as untrustworthy.

Second, our theoretical analysis is conducted under simplified assumptions. Specifically, we mainly analyze the effect of HyperTrust from the perspective of trustworthy message propagation and classification margin. The analysis provides an intuitive explanation of why HyperedgeBoost and HyperedgePrune can improve robustness under label noise. However, practical HGNNs involve nonlinear transformations, learnable parameters, and complex optimization dynamics. Therefore, the current theoretical results should be viewed as an explanatory analysis rather than a complete generalization guarantee for all hypergraph neural networks and all types of label noise.

\end{document}